\documentclass[10pt,twocolumn,letterpaper]{article}
\usepackage[letterpaper,top=0.9in,bottom=0.9in,left=0.75in,right=0.75in,columnsep=0.3in]{geometry}
\usepackage[T1]{fontenc}
\usepackage{amsmath}
\usepackage{newtxtext,newtxmath}
\usepackage{xspace}
\usepackage[dvipsnames]{xcolor}
\usepackage{graphicx}
\usepackage{booktabs}
\usepackage[numbers,sort&compress]{natbib}
\usepackage[font=small,labelfont=bf]{caption}
\usepackage{subcaption}
\usepackage{enumitem}
\usepackage{xurl}
\usepackage{float}
\usepackage{multirow}
\usepackage{makecell}
\usepackage{bm}
\usepackage{float}     %
\usepackage{placeins}   %
\usepackage{pifont}     %

\newif\ifincludesuppl
\includesuppltrue

\newcommand{\method}{TopX\xspace}
\newcommand{\bench}{SkelShift\xspace}

\newcommand{\nSpClips}{635}%
\newcommand{\nSpRigs}{6}%
\newcommand{\nSpRealSkate}{0.213}%
\newcommand{\nSpRealSlide}{1.03}%
\newcommand{\nSpRealCov}{0.199}%
\newcommand{\nSpOursFID}{1.19}%
\newcommand{\nSpOursSkate}{0.208}%
\newcommand{\nSpOursSlide}{1.27}%
\newcommand{\nSpOursCov}{0.234}%
\newcommand{\nSpUniFID}{3.32}%
\newcommand{\nSpUniSkate}{0.268}%
\newcommand{\nSpUniSlide}{1.59}%
\newcommand{\nSpUniCov}{0.220}%
\newcommand{\nSpAnyFID}{24.63}%
\newcommand{\nSpAnySkate}{0.759}%
\newcommand{\nSpAnySlide}{10.65}%
\newcommand{\nSpAnyCov}{0.358}%
\newcommand{\nTbClips}{62}%
\newcommand{\nTbRigs}{6}%
\newcommand{\nTbRealSkate}{0.208}%
\newcommand{\nTbRealSlide}{1.51}%
\newcommand{\nTbRealCov}{0.434}%
\newcommand{\nTbOursFID}{3.05}%
\newcommand{\nTbOursSkate}{0.206}%
\newcommand{\nTbOursSlide}{1.69}%
\newcommand{\nTbOursCov}{0.506}%
\newcommand{\nTbUniFID}{4.62}%
\newcommand{\nTbUniSkate}{0.221}%
\newcommand{\nTbUniSlide}{2.16}%
\newcommand{\nTbUniCov}{0.412}%
\newcommand{\nTbAnyFID}{9.85}%
\newcommand{\nTbAnySkate}{0.271}%
\newcommand{\nTbAnySlide}{4.21}%
\newcommand{\nTbAnyCov}{0.398}%
\newcommand{\nLexExact}{94.2}%
\newcommand{\nLexRigs}{672}%
\newcommand{\nPbIDclipRigs}{93}%
\newcommand{\nPbIDclipClips}{172}%
\newcommand{\nPbIDclipKID}{0.950}%
\newcommand{\nPbIDclipFloor}{0.455}%
\newcommand{\nPbIDclipSlip}{0.309}%
\newcommand{\nPbIDclipSlipReal}{0.353}%
\newcommand{\nPbIDclipPen}{0.204}%
\newcommand{\nPbIDclipFloat}{0.110}%
\newcommand{\nPbIDclipFloatReal}{0.128}%
\newcommand{\nPbIDclipSpeed}{0.87}%
\newcommand{\nPbIDtreeRigs}{504}%
\newcommand{\nPbIDtreeClips}{508}%
\newcommand{\nPbIDtreeKID}{0.145}%
\newcommand{\nPbIDtreeFloor}{0.002}%
\newcommand{\nPbIDtreeSlip}{0.304}%
\newcommand{\nPbIDtreeSlipReal}{0.349}%
\newcommand{\nPbIDtreePen}{0.172}%
\newcommand{\nPbIDtreeFloat}{0.021}%
\newcommand{\nPbIDtreeFloatReal}{0.039}%
\newcommand{\nPbIDtreeSpeed}{0.90}%
\newcommand{\nPbTreeRigs}{218}%
\newcommand{\nPbTreeClips}{320}%
\newcommand{\nPbTreeKID}{1.74}%
\newcommand{\nPbTreeFloor}{0.142}%
\newcommand{\nPbTreeSlip}{0.310}%
\newcommand{\nPbTreeSlipReal}{0.366}%
\newcommand{\nPbTreePen}{0.200}%
\newcommand{\nPbTreeFloat}{0.045}%
\newcommand{\nPbTreeFloatReal}{0.078}%
\newcommand{\nPbTreeSpeed}{0.77}%
\newcommand{\nPbPlanRigs}{45}%
\newcommand{\nPbPlanClips}{62}%
\newcommand{\nPbPlanKID}{10.9}%
\newcommand{\nPbPlanFloor}{6.04}%
\newcommand{\nPbPlanSlip}{0.176}%
\newcommand{\nPbPlanSlipReal}{0.280}%
\newcommand{\nPbPlanPen}{0.251}%
\newcommand{\nPbPlanFloat}{0.106}%
\newcommand{\nPbPlanFloatReal}{0.164}%
\newcommand{\nPbPlanSpeed}{0.81}%
\newcommand{\nPbMixamoRigs}{172}%
\newcommand{\nPbMixamoClips}{350}%
\newcommand{\nPbMixamoKID}{0.664}%
\newcommand{\nPbMixamoFloor}{0.874}%
\newcommand{\nPbMixamoSlip}{0.311}%
\newcommand{\nPbMixamoSlipReal}{0.363}%
\newcommand{\nPbMixamoPen}{0.131}%
\newcommand{\nPbMixamoFloat}{0.024}%
\newcommand{\nPbMixamoFloatReal}{0.050}%
\newcommand{\nPbMixamoSpeed}{0.82}%
\newcommand{\nPbJotherRigs}{351}%
\newcommand{\nPbJotherClips}{512}%
\newcommand{\nPbJotherKID}{0.240}%
\newcommand{\nPbJotherFloor}{0.092}%
\newcommand{\nPbJotherSlip}{0.275}%
\newcommand{\nPbJotherSlipReal}{0.333}%
\newcommand{\nPbJotherPen}{0.162}%
\newcommand{\nPbJotherFloat}{0.021}%
\newcommand{\nPbJotherFloatReal}{0.029}%
\newcommand{\nPbJotherSpeed}{0.77}%
\newcommand{\nPbSpeciesRigs}{6}%
\newcommand{\nPbSpeciesClips}{635}%
\newcommand{\nPbSpeciesKID}{0.127}%
\newcommand{\nPbSpeciesFloor}{0.124}%
\newcommand{\nPbSpeciesSlip}{0.226}%
\newcommand{\nPbSpeciesSlipReal}{0.262}%
\newcommand{\nPbSpeciesPen}{0.061}%
\newcommand{\nPbSpeciesFloat}{0.011}%
\newcommand{\nPbSpeciesFloatReal}{0.012}%
\newcommand{\nPbSpeciesSpeed}{0.93}%
\newcommand{\nPbNewSpRigs}{9}%
\newcommand{\nPbNewSpClips}{47}%
\newcommand{\nPbNewSpKID}{91.5}%
\newcommand{\nPbNewSpFloor}{104}%
\newcommand{\nPbNewSpSlip}{0.023}%
\newcommand{\nPbNewSpSlipReal}{0.222}%
\newcommand{\nPbNewSpPen}{0.240}%
\newcommand{\nPbNewSpFloat}{0.000}%
\newcommand{\nPbNewSpFloatReal}{0.323}%
\newcommand{\nPbNewSpSpeed}{0.19}%
\newcommand{\nPbRigRigs}{6}%
\newcommand{\nPbRigClips}{62}%
\newcommand{\nPbRigKID}{1.08}%
\newcommand{\nPbRigFloor}{0.914}%
\newcommand{\nPbRigSlip}{0.181}%
\newcommand{\nPbRigSlipReal}{0.196}%
\newcommand{\nPbRigPen}{0.290}%
\newcommand{\nPbRigFloat}{0.011}%
\newcommand{\nPbRigFloatReal}{0.017}%
\newcommand{\nPbRigSpeed}{0.94}%
\newcommand{\nPbAllRigs}{1,405}%
\newcommand{\nPbAllClips}{2,669}%
\newcommand{\nPbAllKID}{0.324}%
\newcommand{\nPbAllFloor}{0.129}%
\newcommand{\nPbAllSlip}{0.293}%
\newcommand{\nPbAllSlipReal}{0.345}%
\newcommand{\nPbAllPen}{0.174}%
\newcommand{\nPbAllFloat}{0.033}%
\newcommand{\nPbAllFloatReal}{0.055}%
\newcommand{\nPbAllSpeed}{0.83}%
\newcommand{\nCfgAFID}{1.88}%
\newcommand{\nCfgASkate}{0.231}%
\newcommand{\nCfgASlide}{1.60}%
\newcommand{\nCfgACov}{0.452}%
\newcommand{\nCfgAPen}{0.179}%
\newcommand{\nCfgAFloat}{0.025}%
\newcommand{\nCfgASpeed}{0.78}%
\newcommand{\nCfgAKID}{0.149}%
\newcommand{\nCfgBFID}{1.71}%
\newcommand{\nCfgBSkate}{0.243}%
\newcommand{\nCfgBSlide}{1.74}%
\newcommand{\nCfgBCov}{0.441}%
\newcommand{\nCfgBPen}{0.170}%
\newcommand{\nCfgBFloat}{0.030}%
\newcommand{\nCfgBSpeed}{0.84}%
\newcommand{\nCfgBKID}{0.142}%
\newcommand{\nCfgCFID}{1.79}%
\newcommand{\nCfgCSkate}{0.260}%
\newcommand{\nCfgCSlide}{2.02}%
\newcommand{\nCfgCCov}{0.417}%
\newcommand{\nCfgCPen}{0.172}%
\newcommand{\nCfgCFloat}{0.039}%
\newcommand{\nCfgCSpeed}{0.95}%
\newcommand{\nCfgCKID}{0.176}%
\newcommand{\nCfgRealSkate}{0.254}%
\newcommand{\nCfgRealSlide}{1.71}%
\newcommand{\nCfgRealCov}{0.442}%
\newcommand{\nCfgRealPen}{0.000}%
\newcommand{\nCfgRealFloat}{0.050}%
\newcommand{\nCfgRealFloor}{0.091}%
\newcommand{\nCfgClips}{1,563}%
\newcommand{\nCfgRigs}{638}%
\newcommand{\nUcfgClips}{528}%
\newcommand{\nUcfgAKID}{0.253}%
\newcommand{\nUcfgBKID}{0.302}%
\newcommand{\nUcfgCKID}{0.537}%
\newcommand{\nAbCtlTreeKID}{1.78}%
\newcommand{\nAbCtlTreeKIDhw}{0.03}%
\newcommand{\nAbCtlPlanKID}{11.1}%
\newcommand{\nAbCtlPlanKIDhw}{0.2}%
\newcommand{\nAbCtlNewSpKID}{82.4}%
\newcommand{\nAbCtlNewSpKIDhw}{0.8}%
\newcommand{\nAbCtlRigKID}{1.01}%
\newcommand{\nAbCtlRigKIDhw}{0.11}%
\newcommand{\nAbTroTreeKID}{1.88}%
\newcommand{\nAbTroTreeKIDhw}{0.03}%
\newcommand{\nAbTroPlanKID}{11.7}%
\newcommand{\nAbTroPlanKIDhw}{0.1}%
\newcommand{\nAbTroNewSpKID}{84.4}%
\newcommand{\nAbTroNewSpKIDhw}{1.0}%
\newcommand{\nAbTroRigKID}{1.59}%
\newcommand{\nAbTroRigKIDhw}{0.04}%
\newcommand{\nAbEigTreeKID}{1.83}%
\newcommand{\nAbEigTreeKIDhw}{0.03}%
\newcommand{\nAbEigPlanKID}{11.5}%
\newcommand{\nAbEigPlanKIDhw}{0.2}%
\newcommand{\nAbEigNewSpKID}{83.6}%
\newcommand{\nAbEigNewSpKIDhw}{1.3}%
\newcommand{\nAbEigRigKID}{1.17}%
\newcommand{\nAbEigRigKIDhw}{0.15}%
\newcommand{\nAbAugTreeKID}{1.95}%
\newcommand{\nAbAugTreeKIDhw}{0.04}%
\newcommand{\nAbAugPlanKID}{12.1}%
\newcommand{\nAbAugPlanKIDhw}{0.2}%
\newcommand{\nAbAugNewSpKID}{91.8}%
\newcommand{\nAbAugNewSpKIDhw}{1.8}%
\newcommand{\nAbAugRigKID}{1.03}%
\newcommand{\nAbAugRigKIDhw}{0.03}%
\newcommand{\nAbJtabTreeKID}{1.87}%
\newcommand{\nAbJtabTreeKIDhw}{0.04}%
\newcommand{\nAbJtabPlanKID}{11.5}%
\newcommand{\nAbJtabPlanKIDhw}{0.3}%
\newcommand{\nAbJtabNewSpKID}{83.7}%
\newcommand{\nAbJtabNewSpKIDhw}{0.7}%
\newcommand{\nAbJtabRigKID}{1.12}%
\newcommand{\nAbJtabRigKIDhw}{0.11}%
\newcommand{\nAbTreeFloor}{0.142}%
\newcommand{\nAbTreeClips}{320}%
\newcommand{\nAbTreeRigs}{218}%
\newcommand{\nAbPlanFloor}{6.04}%
\newcommand{\nAbPlanClips}{62}%
\newcommand{\nAbPlanRigs}{45}%
\newcommand{\nAbNewSpFloor}{104}%
\newcommand{\nAbNewSpClips}{47}%
\newcommand{\nAbNewSpRigs}{9}%
\newcommand{\nAbRigFloor}{0.914}%
\newcommand{\nAbRigClips}{62}%
\newcommand{\nAbRigRigs}{6}%
\newcommand{\nRcClips}{398}%
\newcommand{\nRcRigs}{196}%
\newcommand{\nRcDeg}{30}%
\newcommand{\nRcBuckets}{7}%
\newcommand{\nRcCtl}{0.44}%
\newcommand{\nRcSeedCtl}{1.49}%
\newcommand{\nRcAug}{0.89}%
\newcommand{\nRcSeedAug}{1.36}%
\newcommand{\nRcDiff}{0.45}%
\newcommand{\nRcLo}{0.41}%
\newcommand{\nRcHi}{0.50}%
\newcommand{\nRcRigLo}{0.39}%
\newcommand{\nRcRigHi}{0.51}%
\newcommand{\nRetOurs}{0.953}%
\newcommand{\nRetReal}{0.929}%
\newcommand{\nRetClips}{439}%
\newcommand{\nRetRigs}{6}%
\newcommand{\nTwoAllClips}{1,062}%
\newcommand{\nTwoAllDescOurs}{4.83}%
\newcommand{\nTwoAllDescUni}{6.06}%
\newcommand{\nTwoAllDescOursB}{\textbf{4.83}}%
\newcommand{\nTwoAllDescUniB}{6.06}%
\newcommand{\nTwoAllKinOurs}{2.00}%
\newcommand{\nTwoAllKinUni}{2.05}%
\newcommand{\nTwoAllKinOursB}{2.00}%
\newcommand{\nTwoAllKinUniB}{2.05}%
\newcommand{\nTwoAllBkOursB}{1.85}%
\newcommand{\nTwoAllBkUniB}{1.90}%
\newcommand{\nTwoAllPenOurs}{0.24}%
\newcommand{\nTwoAllPenUni}{0.33}%
\newcommand{\nTwoAllPenOursB}{\textbf{0.24}}%
\newcommand{\nTwoAllPenUniB}{0.33}%
\newcommand{\nTwoAllSkateReal}{0.35}%
\newcommand{\nTwoAllSkateOursB}{0.33}%
\newcommand{\nTwoAllSkateUniB}{0.35}%
\newcommand{\nTwoAllCovReal}{0.64}%
\newcommand{\nTwoAllCovOursB}{0.60}%
\newcommand{\nTwoAllCovUniB}{0.58}%
\newcommand{\nTwoAllSpeedOurs}{0.93}%
\newcommand{\nTwoAllSpeedUni}{1.08}%
\newcommand{\nTwoAllDescFloor}{0.48}%
\newcommand{\nTwoNoTrClips}{890}%
\newcommand{\nTwoNoTrDescOursB}{\textbf{4.58}}%
\newcommand{\nTwoNoTrDescUniB}{5.73}%
\newcommand{\nTwoNoTrKinOursB}{0.55}%
\newcommand{\nTwoNoTrKinUniB}{0.48}%
\newcommand{\nTwoNoTrBkOursB}{0.56}%
\newcommand{\nTwoNoTrBkUniB}{0.44}%
\newcommand{\nTwoNoTrPenOursB}{\textbf{0.24}}%
\newcommand{\nTwoNoTrPenUniB}{0.33}%
\newcommand{\nTwoNoTrSkateReal}{0.35}%
\newcommand{\nTwoNoTrSkateOursB}{0.33}%
\newcommand{\nTwoNoTrSkateUniB}{0.34}%
\newcommand{\nTwoNoTrCovReal}{0.66}%
\newcommand{\nTwoNoTrCovOursB}{0.61}%
\newcommand{\nTwoNoTrCovUniB}{0.59}%
\newcommand{\nTwoNoTrSpeedOurs}{0.92}%
\newcommand{\nTwoNoTrSpeedUni}{1.06}%
\newcommand{\nTwoNoTrDescFloor}{0.55}%
\newcommand{\nTwoNovelClips}{453}%
\newcommand{\nTwoNovelDescOursB}{\textbf{9.63}}%
\newcommand{\nTwoNovelDescUniB}{10.70}%
\newcommand{\nTwoNovelKinOursB}{1.15}%
\newcommand{\nTwoNovelKinUniB}{0.89}%
\newcommand{\nTwoNovelBkOursB}{1.05}%
\newcommand{\nTwoNovelBkUniB}{0.77}%
\newcommand{\nTwoNovelPenOursB}{\textbf{0.27}}%
\newcommand{\nTwoNovelPenUniB}{0.35}%
\newcommand{\nTwoNovelSkateReal}{0.30}%
\newcommand{\nTwoNovelSkateOursB}{0.26}%
\newcommand{\nTwoNovelSkateUniB}{0.33}%
\newcommand{\nTwoNovelCovReal}{0.57}%
\newcommand{\nTwoNovelCovOursB}{0.52}%
\newcommand{\nTwoNovelCovUniB}{0.47}%
\newcommand{\nTwoNovelSpeedOurs}{0.90}%
\newcommand{\nTwoNovelSpeedUni}{1.13}%
\newcommand{\nTwoNovelDescFloor}{1.13}%
\newcommand{\nTwoReusedClips}{437}%
\newcommand{\nTwoReusedDescOursB}{\textbf{0.65}}%
\newcommand{\nTwoReusedDescUniB}{1.64}%
\newcommand{\nTwoReusedKinOursB}{\textbf{0.06}}%
\newcommand{\nTwoReusedKinUniB}{0.24}%
\newcommand{\nTwoReusedBkOursB}{\textbf{0.08}}%
\newcommand{\nTwoReusedBkUniB}{0.27}%
\newcommand{\nTwoReusedPenOursB}{\textbf{0.23}}%
\newcommand{\nTwoReusedPenUniB}{0.32}%
\newcommand{\nTwoReusedSkateReal}{0.39}%
\newcommand{\nTwoReusedSkateOursB}{\textbf{0.39}}%
\newcommand{\nTwoReusedSkateUniB}{0.35}%
\newcommand{\nTwoReusedCovReal}{0.75}%
\newcommand{\nTwoReusedCovOursB}{0.69}%
\newcommand{\nTwoReusedCovUniB}{0.70}%
\newcommand{\nTwoReusedSpeedOurs}{0.94}%
\newcommand{\nTwoReusedSpeedUni}{1.02}%
\newcommand{\nTwoReusedDescFloor}{0.60}%
\newcommand{\nTwoSTreeClips}{508}%
\newcommand{\nTwoSTreeDescOurs}{1.06}%
\newcommand{\nTwoSTreeDescUni}{2.72}%
\newcommand{\nTwoSTreeDescOursB}{\textbf{1.06}}%
\newcommand{\nTwoSTreeDescUniB}{2.72}%
\newcommand{\nTwoSTreeKinOursB}{\textbf{0.04}}%
\newcommand{\nTwoSTreeKinUniB}{0.38}%
\newcommand{\nTwoSTreeBkOursB}{0.06}%
\newcommand{\nTwoSTreeBkUniB}{0.47}%
\newcommand{\nTwoSTreePenOursB}{\textbf{0.24}}%
\newcommand{\nTwoSTreePenUniB}{0.34}%
\newcommand{\nTwoSTreeSkateReal}{0.34}%
\newcommand{\nTwoSTreeSkateOursB}{0.33}%
\newcommand{\nTwoSTreeSkateUniB}{0.33}%
\newcommand{\nTwoSTreeCovReal}{0.74}%
\newcommand{\nTwoSTreeCovOursB}{0.70}%
\newcommand{\nTwoSTreeCovUniB}{0.69}%
\newcommand{\nTwoSTreeSpeedOurs}{0.98}%
\newcommand{\nTwoSTreeSpeedUni}{1.08}%
\newcommand{\nTwoSTreeDescFloor}{0.69}%
\newcommand{\nTwoTreeClips}{320}%
\newcommand{\nTwoTreeDescOursB}{8.38}%
\newcommand{\nTwoTreeDescUniB}{9.02}%
\newcommand{\nTwoTreeKinOursB}{1.33}%
\newcommand{\nTwoTreeKinUniB}{1.24}%
\newcommand{\nTwoTreeBkOursB}{1.18}%
\newcommand{\nTwoTreeBkUniB}{1.12}%
\newcommand{\nTwoTreePenOursB}{\textbf{0.22}}%
\newcommand{\nTwoTreePenUniB}{0.30}%
\newcommand{\nTwoTreeSkateReal}{0.38}%
\newcommand{\nTwoTreeSkateOursB}{0.34}%
\newcommand{\nTwoTreeSkateUniB}{0.39}%
\newcommand{\nTwoTreeCovReal}{0.57}%
\newcommand{\nTwoTreeCovOursB}{0.50}%
\newcommand{\nTwoTreeCovUniB}{0.46}%
\newcommand{\nTwoTreeSpeedOurs}{0.82}%
\newcommand{\nTwoTreeSpeedUni}{1.04}%
\newcommand{\nTwoTreeDescFloor}{1.39}%
\newcommand{\nTwoPlanClips}{62}%
\newcommand{\nTwoPlanDescOursB}{24.1}%
\newcommand{\nTwoPlanDescUniB}{22.3}%
\newcommand{\nTwoPlanKinOursB}{3.02}%
\newcommand{\nTwoPlanKinUniB}{2.65}%
\newcommand{\nTwoPlanBkOursB}{2.32}%
\newcommand{\nTwoPlanBkUniB}{1.90}%
\newcommand{\nTwoPlanPenOursB}{0.35}%
\newcommand{\nTwoPlanPenUniB}{0.35}%
\newcommand{\nTwoPlanSkateReal}{0.19}%
\newcommand{\nTwoPlanSkateOursB}{0.12}%
\newcommand{\nTwoPlanSkateUniB}{0.23}%
\newcommand{\nTwoPlanCovReal}{0.21}%
\newcommand{\nTwoPlanCovOursB}{0.13}%
\newcommand{\nTwoPlanCovUniB}{0.09}%
\newcommand{\nTwoPlanSpeedOurs}{0.65}%
\newcommand{\nTwoPlanSpeedUni}{0.81}%
\newcommand{\nTwoClipClips}{172}%
\newcommand{\nTwoClipDescOursB}{\textbf{7.18}}%
\newcommand{\nTwoClipDescUniB}{9.09}%
\newcommand{\nTwoClipKinOursB}{2.40}%
\newcommand{\nTwoClipKinUniB}{2.49}%
\newcommand{\nTwoClipBkOursB}{2.10}%
\newcommand{\nTwoClipBkUniB}{2.14}%
\newcommand{\nTwoClipPenOursB}{\textbf{0.25}}%
\newcommand{\nTwoClipPenUniB}{0.34}%
\newcommand{\nTwoClipSkateReal}{0.33}%
\newcommand{\nTwoClipSkateOursB}{0.34}%
\newcommand{\nTwoClipSkateUniB}{0.37}%
\newcommand{\nTwoClipCovReal}{0.55}%
\newcommand{\nTwoClipCovOursB}{0.52}%
\newcommand{\nTwoClipCovUniB}{0.51}%
\newcommand{\nTwoClipSpeedOurs}{0.98}%
\newcommand{\nTwoClipSpeedUni}{1.23}%
\newcommand{\nTwoClipDescFloor}{2.83}%
\newcommand{\nThrAllClips}{1,062}%
\newcommand{\nThrAllDescUni}{4.32}%
\newcommand{\nThrAllDescOursB}{4.83}%
\newcommand{\nThrAllDescUniB}{\textbf{4.32}}%
\newcommand{\nThrAllKinUni}{2.52}%
\newcommand{\nThrAllKinOursB}{\textbf{2.00}}%
\newcommand{\nThrAllKinUniB}{2.52}%
\newcommand{\nThrAllBkOursB}{\textbf{1.85}}%
\newcommand{\nThrAllBkUniB}{2.18}%
\newcommand{\nThrAllPenUni}{0.28}%
\newcommand{\nThrAllPenOursB}{\textbf{0.24}}%
\newcommand{\nThrAllPenUniB}{0.28}%
\newcommand{\nThrAllSkateReal}{0.35}%
\newcommand{\nThrAllSkateOursB}{\textbf{0.33}}%
\newcommand{\nThrAllSkateUniB}{0.24}%
\newcommand{\nThrAllSpeedOursB}{\textbf{0.93}}%
\newcommand{\nThrAllSpeedUniB}{0.85}%
\newcommand{\nThrNoTrClips}{890}%
\newcommand{\nThrNoTrDescOursB}{4.58}%
\newcommand{\nThrNoTrDescUniB}{4.22}%
\newcommand{\nThrNoTrKinOursB}{\textbf{0.55}}%
\newcommand{\nThrNoTrKinUniB}{1.48}%
\newcommand{\nThrNoTrBkOursB}{\textbf{0.56}}%
\newcommand{\nThrNoTrBkUniB}{1.13}%
\newcommand{\nThrNoTrPenOursB}{\textbf{0.24}}%
\newcommand{\nThrNoTrPenUniB}{0.28}%
\newcommand{\nThrNoTrSkateReal}{0.35}%
\newcommand{\nThrNoTrSkateOursB}{\textbf{0.33}}%
\newcommand{\nThrNoTrSkateUniB}{0.24}%
\newcommand{\nThrNoTrSpeedOursB}{\textbf{0.92}}%
\newcommand{\nThrNoTrSpeedUniB}{0.83}%
\newcommand{\nThrNovelClips}{453}%
\newcommand{\nThrNovelDescOursB}{9.63}%
\newcommand{\nThrNovelDescUniB}{\textbf{7.95}}%
\newcommand{\nThrNovelKinOursB}{\textbf{1.15}}%
\newcommand{\nThrNovelKinUniB}{2.17}%
\newcommand{\nThrNovelBkOursB}{\textbf{1.05}}%
\newcommand{\nThrNovelBkUniB}{1.70}%
\newcommand{\nThrNovelPenOursB}{0.27}%
\newcommand{\nThrNovelPenUniB}{0.29}%
\newcommand{\nThrNovelSkateReal}{0.30}%
\newcommand{\nThrNovelSkateOursB}{\textbf{0.26}}%
\newcommand{\nThrNovelSkateUniB}{0.21}%
\newcommand{\nThrNovelSpeedOursB}{\textbf{0.90}}%
\newcommand{\nThrNovelSpeedUniB}{0.83}%
\newcommand{\nThrReusedClips}{437}%
\newcommand{\nThrReusedDescOursB}{\textbf{0.65}}%
\newcommand{\nThrReusedDescUniB}{1.21}%
\newcommand{\nThrReusedKinOursB}{\textbf{0.06}}%
\newcommand{\nThrReusedKinUniB}{0.72}%
\newcommand{\nThrReusedBkOursB}{\textbf{0.08}}%
\newcommand{\nThrReusedBkUniB}{0.47}%
\newcommand{\nThrReusedPenOursB}{\textbf{0.23}}%
\newcommand{\nThrReusedPenUniB}{0.27}%
\newcommand{\nThrReusedSkateReal}{0.39}%
\newcommand{\nThrReusedSkateOursB}{\textbf{0.39}}%
\newcommand{\nThrReusedSkateUniB}{0.27}%
\newcommand{\nThrReusedSpeedOursB}{\textbf{0.94}}%
\newcommand{\nThrReusedSpeedUniB}{0.83}%
\newcommand{\nThrSTreeClips}{508}%
\newcommand{\nThrSTreeDescUni}{1.56}%
\newcommand{\nThrSTreeDescOursB}{\textbf{1.06}}%
\newcommand{\nThrSTreeDescUniB}{1.56}%
\newcommand{\nThrSTreeKinOursB}{\textbf{0.04}}%
\newcommand{\nThrSTreeKinUniB}{0.53}%
\newcommand{\nThrSTreeBkOursB}{\textbf{0.06}}%
\newcommand{\nThrSTreeBkUniB}{0.34}%
\newcommand{\nThrSTreePenOursB}{\textbf{0.24}}%
\newcommand{\nThrSTreePenUniB}{0.29}%
\newcommand{\nThrSTreeSkateReal}{0.34}%
\newcommand{\nThrSTreeSkateOursB}{\textbf{0.33}}%
\newcommand{\nThrSTreeSkateUniB}{0.24}%
\newcommand{\nThrSTreeSpeedOursB}{\textbf{0.98}}%
\newcommand{\nThrSTreeSpeedUniB}{0.88}%
\newcommand{\nThrTreeClips}{320}%
\newcommand{\nThrTreeDescOursB}{8.38}%
\newcommand{\nThrTreeDescUniB}{\textbf{7.17}}%
\newcommand{\nThrTreeKinOursB}{\textbf{1.33}}%
\newcommand{\nThrTreeKinUniB}{2.63}%
\newcommand{\nThrTreeBkOursB}{\textbf{1.18}}%
\newcommand{\nThrTreeBkUniB}{2.13}%
\newcommand{\nThrTreePenOursB}{0.22}%
\newcommand{\nThrTreePenUniB}{0.26}%
\newcommand{\nThrTreeSkateReal}{0.38}%
\newcommand{\nThrTreeSkateOursB}{\textbf{0.34}}%
\newcommand{\nThrTreeSkateUniB}{0.25}%
\newcommand{\nThrTreeSpeedOursB}{0.82}%
\newcommand{\nThrTreeSpeedUniB}{0.77}%
\newcommand{\nThrPlanClips}{62}%
\newcommand{\nThrPlanDescOursB}{24.1}%
\newcommand{\nThrPlanDescUniB}{20.4}%
\newcommand{\nThrPlanKinOursB}{3.02}%
\newcommand{\nThrPlanKinUniB}{5.46}%
\newcommand{\nThrPlanBkOursB}{2.32}%
\newcommand{\nThrPlanBkUniB}{3.70}%
\newcommand{\nThrPlanPenOursB}{0.35}%
\newcommand{\nThrPlanPenUniB}{0.27}%
\newcommand{\nThrPlanSkateReal}{0.19}%
\newcommand{\nThrPlanSkateOursB}{0.12}%
\newcommand{\nThrPlanSkateUniB}{0.07}%
\newcommand{\nThrPlanSpeedOursB}{\textbf{0.65}}%
\newcommand{\nThrPlanSpeedUniB}{0.45}%
\newcommand{\nThrClipClips}{172}%
\newcommand{\nThrClipDescOursB}{7.18}%
\newcommand{\nThrClipDescUniB}{5.96}%
\newcommand{\nThrClipKinOursB}{2.40}%
\newcommand{\nThrClipKinUniB}{2.57}%
\newcommand{\nThrClipBkOursB}{2.10}%
\newcommand{\nThrClipBkUniB}{2.22}%
\newcommand{\nThrClipPenOursB}{0.25}%
\newcommand{\nThrClipPenUniB}{0.27}%
\newcommand{\nThrClipSkateReal}{0.33}%
\newcommand{\nThrClipSkateOursB}{0.34}%
\newcommand{\nThrClipSkateUniB}{0.25}%
\newcommand{\nThrClipSpeedOursB}{0.98}%
\newcommand{\nThrClipSpeedUniB}{0.93}%
\newcommand{\nTwoOursSteps}{200}%
\newcommand{\nTwoOursCfg}{2}%
\newcommand{\nTwoUniCfg}{7}%
\newcommand{\nTwoDevUniOnePointFive}{2.32}%
\newcommand{\nThrDevUniDescOnePointFive}{4.15}%
\newcommand{\nTwoDevUniTwo}{2.21}%
\newcommand{\nThrDevUniDescTwo}{3.92}%
\newcommand{\nTwoDevUniThree}{1.96}%
\newcommand{\nThrDevUniDescThree}{3.63}%
\newcommand{\nTwoDevUniFour}{1.83}%
\newcommand{\nThrDevUniDescFour}{3.73}%
\newcommand{\nTwoDevUniFive}{1.72}%
\newcommand{\nThrDevUniDescFive}{4.10}%
\newcommand{\nTwoDevUniSeven}{1.65}%
\newcommand{\nThrDevUniDescSeven}{5.24}%
\newcommand{\nTwoDevUniTen}{1.68}%
\newcommand{\nThrDevUniDescTen}{7.19}%
\newcommand{\nThrUniCfg}{3}%
\newcommand{\nTwoDevOursTwoStepsTwoHundred}{1.34}%
\newcommand{\nTwoDevOursTwoStepsFourHundred}{1.29}%
\newcommand{\nTwoDevOursDescOnePointFive}{3.87}%
\newcommand{\nTwoDevOursDescTwo}{3.66}%
\newcommand{\nTwoDevOursDescThree}{3.87}%
\newcommand{\nCostOursClipTwentyNfe}{40}%
\newcommand{\nCostOursClipTwentySec}{0.67}%
\newcommand{\nCostOursClipFiftyNfe}{100}%
\newcommand{\nCostOursClipFiftySec}{1.63}%
\newcommand{\nCostOursClipHundredNfe}{200}%
\newcommand{\nCostOursClipHundredSec}{3.23}%
\newcommand{\nCostOursClipTwoHundredNfe}{400}%
\newcommand{\nCostOursClipTwoHundredSec}{6.43}%
\newcommand{\nCostOursClipFourHundredNfe}{800}%
\newcommand{\nCostOursClipFourHundredSec}{12.82}%
\newcommand{\nCostUniThreeNfe}{366.4}%
\newcommand{\nCostUniThreeSec}{6.13}%
\newcommand{\nCostUniFourNfe}{382.5}%
\newcommand{\nCostUniFourSec}{6.32}%
\newcommand{\nCostUniFiveNfe}{398.3}%
\newcommand{\nCostUniFiveSec}{6.73}%
\newcommand{\nCostUniSevenNfe}{414.6}%
\newcommand{\nCostUniSevenSec}{6.85}%
\newcommand{\nCostUniSevenNfeBatch}{548.6}%
\newcommand{\nTaRealSixtyAllR}{0.896}%
\newcommand{\nTaRealFullAllR}{0.929}%
\newcommand{\nTaMainSixtyAllR}{0.915}%
\newcommand{\nTaMainSixtyAllRHw}{0.007}%
\newcommand{\nTaMainSixtyAllShuf}{0.122}%
\newcommand{\nTaMainClipSixtyAllR}{0.932}%
\newcommand{\nTaMainClipSixtyAllRHw}{0.007}%
\newcommand{\nTaMainClipSixtyAllShuf}{0.113}%
\newcommand{\nTaMainClipFullAllR}{0.953}%
\newcommand{\nTaMainClipFullAllRHw}{0.004}%
\newcommand{\nTaMainClipFullAllShuf}{0.120}%
\newcommand{\nTaUniOffAllR}{0.279}%
\newcommand{\nTaUniOffAllRHw}{0.009}%
\newcommand{\nTaUniOffAllShuf}{0.130}%
\newcommand{\nTaSameAllR}{0.255}%
\newcommand{\nTaSameAllRHw}{0.008}%
\newcommand{\nTaSameAllShuf}{0.118}%
\newcommand{\nTaAllClips}{439}%
\newcommand{\nTaAllRigs}{6}%
\newcommand{\nTaRealSixtyEmuR}{0.871}%
\newcommand{\nTaRealFullEmuR}{0.933}%
\newcommand{\nTaMainSixtyEmuR}{0.928}%
\newcommand{\nTaMainSixtyEmuRHw}{0.006}%
\newcommand{\nTaMainSixtyEmuShuf}{0.137}%
\newcommand{\nTaMainClipSixtyEmuR}{0.922}%
\newcommand{\nTaMainClipSixtyEmuRHw}{0.016}%
\newcommand{\nTaMainClipSixtyEmuShuf}{0.127}%
\newcommand{\nTaMainClipFullEmuR}{0.954}%
\newcommand{\nTaMainClipFullEmuRHw}{0.013}%
\newcommand{\nTaMainClipFullEmuShuf}{0.137}%
\newcommand{\nTaUniOffEmuR}{0.337}%
\newcommand{\nTaUniOffEmuRHw}{0.015}%
\newcommand{\nTaUniOffEmuShuf}{0.138}%
\newcommand{\nTaSameEmuR}{0.317}%
\newcommand{\nTaSameEmuRHw}{0.019}%
\newcommand{\nTaSameEmuShuf}{0.142}%
\newcommand{\nTaUniReEmuR}{0.192}%
\newcommand{\nTaUniReEmuRHw}{0.007}%
\newcommand{\nTaUniReEmuShuf}{0.147}%
\newcommand{\nTaEmuClips}{181}%
\newcommand{\nTaEmuRigs}{3}%

\newcommand{\R}{\mathbb{R}}
\newcommand{\x}{\mathbf{x}}
\newcommand{\Prest}{\mathbf{P}^{\mathrm{rest}}}
\newcommand{\fidk}{FID$_{\mathrm{kin}}$\xspace}      %
\newcommand{\kidk}{KID$_{\mathrm{desc}}$\xspace}     %
\newcommand{\fidd}{FID$_{\mathrm{desc}}$\xspace}     %

\newcommand{\ie}{i.e.\@\xspace}

\usepackage[breaklinks,colorlinks,linkcolor=BrickRed,citecolor=OliveGreen,urlcolor=NavyBlue]{hyperref}
\usepackage[capitalize]{cleveref}
\crefname{section}{Sec.}{Secs.}
\Crefname{section}{Section}{Sections}
\crefname{table}{Tab.}{Tabs.}
\Crefname{table}{Table}{Tables}
\crefname{figure}{Fig.}{Figs.}
\Crefname{figure}{Figure}{Figures}
\crefname{equation}{Eq.}{Eqs.}
\Crefname{equation}{Equation}{Equations}
\title{\method{}: Text-to-Motion for Unseen Skeletons from the Skeleton File Alone}
\author{%
Tengjiao Sun$^{1,3}$ \enskip Pengcheng Fang$^{1,3}$ \enskip Xiaoyu Zhan$^{2,3}$ \enskip Dongjie Fu$^{2}$ \enskip Xiaohao Cai$^{1}$ \enskip Hansung Kim$^{1\dagger}$\\[4pt]
{\normalsize $^1$University of Southampton \qquad $^2$Nanjing University \qquad $^3$MOGO AI}\\[2pt]
{\small $^\dagger$Corresponding author}
}
\date{}
\begin{document}
\maketitle
\begin{abstract}
Text-to-motion models generate motion only for the skeletons they were trained on, while a new character arrives as a skeleton file
and nothing else. We study text-to-motion for skeletons unseen in training, without clips of the character, templates, retargeting or
fine-tuning. We present \method, a 73M-parameter flow-matching
transformer that generates joint rotations and a root trajectory for any skeleton from its kinematic tree, rest pose and joint names. Motion is normalised by the rest pose and the size of the skeleton, and the rest pose is encoded as a world-frame descriptor that is independent
of how the rig's author oriented each joint, while rest conventions are randomised during training. Attention is factorised over the joints and the frames of a motion. Across joints, queries and keys are rotated by HK-RoPE,
a heat-kernel rotary encoding of the skeleton graph that is defined for every tree and does not change with the order of
its joints or the eigenbasis of its Laplacian; across frames, they are rotated by a rotary encoding of frame order. To evaluate this generalisation, we build \bench, which unifies four motion corpora under one codec and holds out whole
skeletons, tree classes and species, so that each test bucket isolates one kind of skeleton shift: topology, joint count,
body plan, species or rig source. It scores motion quality without a learned evaluator. A single model animates every benchmark skeleton (4--117 joints) directly from its skeleton file and generates up to
240 frames. On held-out game species its motion is about as close to the real clips as two halves
of the real clips are to each other (kernel distance \nPbSpeciesKID{} against \nPbSpeciesFloor), and text-to-motion retrieval reaches R@1 \nRetOurs, compared with \nRetReal{} for real motion.
Project page: \url{https://chdtevior.github.io/TopX-page/}.
\end{abstract}

\section{Introduction}\label{sec:intro}

Text-to-motion generation has become highly capable for the human body~\cite{guo2022generating,tevet2023mdm,guo2024momask,fan2025motionmillion},
and nearly all existing methods assume a single skeleton: the joint count, joint order and rest pose of the training body are built into the
model. Games, film and virtual production animate large casts of non-human characters, each rigged with its own kinematic tree,
joint names, joint order and rest pose. When such a character is created, only its skeleton file is available: there are no clips to normalise with, adapt to or
retarget from. A text-to-motion model for such pipelines must therefore animate the skeleton from this file alone.

Three properties of a skeleton file make this challenging. \emph{Scale}: without clips, the size of a character's motion and the
spread of each joint's channels are unknown, so normalisation by per-rig motion statistics cannot be applied.
\emph{Conventions}: the order in which a file lists its joints, the names it gives them and the orientation of each joint's
rest frame are arbitrary choices of the rig's author, on which the generated motion should not
depend. \emph{Shape}: the tree itself may lie outside the training distribution, with a new topology, more joints or an unfamiliar body plan. Existing approaches address these only in part: retargeting~\cite{villegas2018nkn,aberman2020skeleton} and template-based
models~\cite{yang2024omnimotiongpt,wang2025xmogen,wang2025animo} cannot represent skeletons beyond their template, and
AnyTop~\cite{gat2025anytop} requires statistics of the target skeleton's own motions.

The extent to which such models generalise remains unclear, because prior evaluations on unseen skeletons use only a
handful of held-out skeleton types~\cite{mou2026unimate}, held-out species sharing one topology~\cite{wang2025xmogen},
or no text conditioning~\cite{pu2026topocap}. Measuring generalisation requires held-out skeletons whose distance from the training data is
controlled and reported, and metrics that remain valid on them; since the accuracy that a learned text--motion evaluator attains even on real
motion can drop substantially outside its training distribution~\cite{wang2025xmogen}, we score motion quality with evaluator-free statistics.

To address these challenges, we make three contributions:
\begin{itemize}
\item \textbf{\method, a text-to-motion system for unseen skeletons.} \method{} is a 73M-parameter flow-matching
  transformer that animates any skeleton from its skeleton file alone (kinematic tree, rest pose and joint names), without
  clips of the target, retargeting or fine-tuning. Normalising motion by the rest pose handles \emph{scale} without
  per-rig statistics, and the rest pose is described by a world-frame descriptor that is independent of the rig's
  rest-frame \emph{conventions}; the model outputs joint rotations and a root trajectory that drive the rig directly
  (\cref{fig:arch}, \cref{sec:method}).
\item \textbf{HK-RoPE, a graph positional encoding for skeletons of arbitrary topology.} To handle unseen tree
  \emph{shapes}, HK-RoPE rotates attention queries and keys by heat-kernel coordinates of the skeleton graph, which are
  defined for every tree, invariant to joint relabelling and independent of the Laplacian eigenbasis
  (\cref{sec:method:model}).
\item \textbf{\bench, a dataset and benchmark for unseen skeletons.} To make generalisation measurable, the dataset
  unifies four motion corpora under one codec: 143,334 captioned clips on skeletons of 4 to 142 joints, with every joint
  described in text. The benchmark holds out whole skeletons, tree classes and species in buckets along five axes of
  skeleton shift, reports each test tree's distance to the training trees, and scores motion quality without a learned
  evaluator (\cref{sec:bench}).
\end{itemize}

\section{Related Work}\label{sec:related}

\paragraph{Text-to-motion.} Human text-to-motion has progressed from variational models~\cite{guo2022generating} to
diffusion in pose space~\cite{tevet2023mdm,zhang2024motiondiffuse} or latent space~\cite{chen2023mld} and to token-based
models over quantised motion~\cite{zhang2023t2mgpt,guo2024momask}, and has been extended with larger
corpora~\cite{lin2023motionx,fan2025motionmillion} and skeleton-aware latents~\cite{hong2025salad}. Animal text-to-motion methods rely on a fixed template, such as the SMAL body
model~\cite{yang2024omnimotiongpt}, a topology shared by humans and animals~\cite{wang2025xmogen} or a unified 30-joint
skeleton~\cite{wang2025animo};
OmniSkel adapts a human model to new bone lengths~\cite{liu2025omniskel}. All of these fix the kinematic tree, or at least its topology, at training time. In contrast, \method{} fixes neither: a
single model takes any kinematic tree as input, so a new character can be animated directly from its skeleton file.

\paragraph{Motion on arbitrary skeletons.} Retargeting maps motion between structurally compatible skeletons~\cite{villegas2018nkn,aberman2020skeleton}
or through skeleton-agnostic embeddings~\cite{lee2023same,zhang2026sata}, and Motion2Motion~\cite{chen2025motion2motion}
transfers across topologies from a few examples of the target. AnyTop~\cite{gat2025anytop} generates motion on arbitrary trees without text
conditioning; it places the rest pose as the first frame, biases attention by tree distances, and normalises each skeleton
by statistics of that skeleton's own motions; UniMoGen~\cite{khani2025unimogen} adds style
and trajectory control on human skeletons. Text-conditioned generation across topologies has been
demonstrated by OmniZoo~\cite{chen2025omnizoo}, G-MDM~\cite{lee2025dragon}, which predicts joint rotations without root
translation, and ACT~\cite{long2026act}, which rigs a mesh and animates it from text. Concurrently with this work,
UniMate~\cite{mou2026unimate} animates unseen skeletons zero-shot using a spectral rotary encoding of Laplacian
eigenvectors, a graph attention bias, rest-pose rebasing and skeleton augmentation. \method{} addresses the same setting but differs in four design choices: motion is
normalised by the rest pose and the skeleton's size, not by per-rig motion statistics; the rest pose is encoded by a world-frame descriptor that is independent of how each joint was
oriented, and rest conventions are randomised during training; the joint coordinate of HK-RoPE is independent of the Laplacian eigenbasis, so the output does not change
with joint order even under repeated eigenvalues; and one model generates up to 240 frames on skeletons of 4 to 117 joints. 

\paragraph{Evaluating unseen skeletons.} UniMate holds out seven Truebones skeleton types (58 sequences) and reports a kinematic-feature FID and diversity, as well as foot sliding on 20 legged skeletons; X-MoGen~\cite{wang2025xmogen} holds out species on its shared topology. For
video-driven animation, TopoCap~\cite{pu2026topocap} separates seen and unseen topologies and
MoCapAnything~\cite{gong2026mocapanything} stratifies seen, rare and unseen species; in retargeting, Hong et al.~\cite{hong2026skinned} report proportion changes and arbitrary structures separately. \bench{} brings such controlled axes of evaluation to text-conditioned generation. It tests whether a single model
generalises to unseen topologies, rare body plans, joint counts outside the training range, new species and held-out Truebones rigs, and on each axis it measures how realistic the generated motion is, how well it keeps ground contact and,
on held-out species, how well it matches the text. Held-out units are removed from training, the distance from every test tree to the training trees is
reported, and results are broken down by axis.

\paragraph{Positional encodings on graphs.} Laplacian eigenvectors are ambiguous in sign and, within the eigenspace of a
repeated eigenvalue, in the choice of basis~\cite{lim2023signnet,huang2024spe}. Diffusion-based coordinates avoid both: random-walk~\cite{dwivedi2022rwpe} and
heat-kernel~\cite{sun2009hks} signatures are well-established node encodings~\cite{rampasek2022graphgps,canturk2024gpse}, and
WIRE~\cite{reid2026wire} uses spectral and random-walk coordinates as rotary angles~\cite{su2024roformer}. HK-RoPE adopts the heat-kernel signature as the rotary coordinate of a skeleton's joints
and pairs it with a rotary encoding of frame order.

\section{Method}\label{sec:method}

\begin{figure*}[t]
\centering
\includegraphics[width=\linewidth]{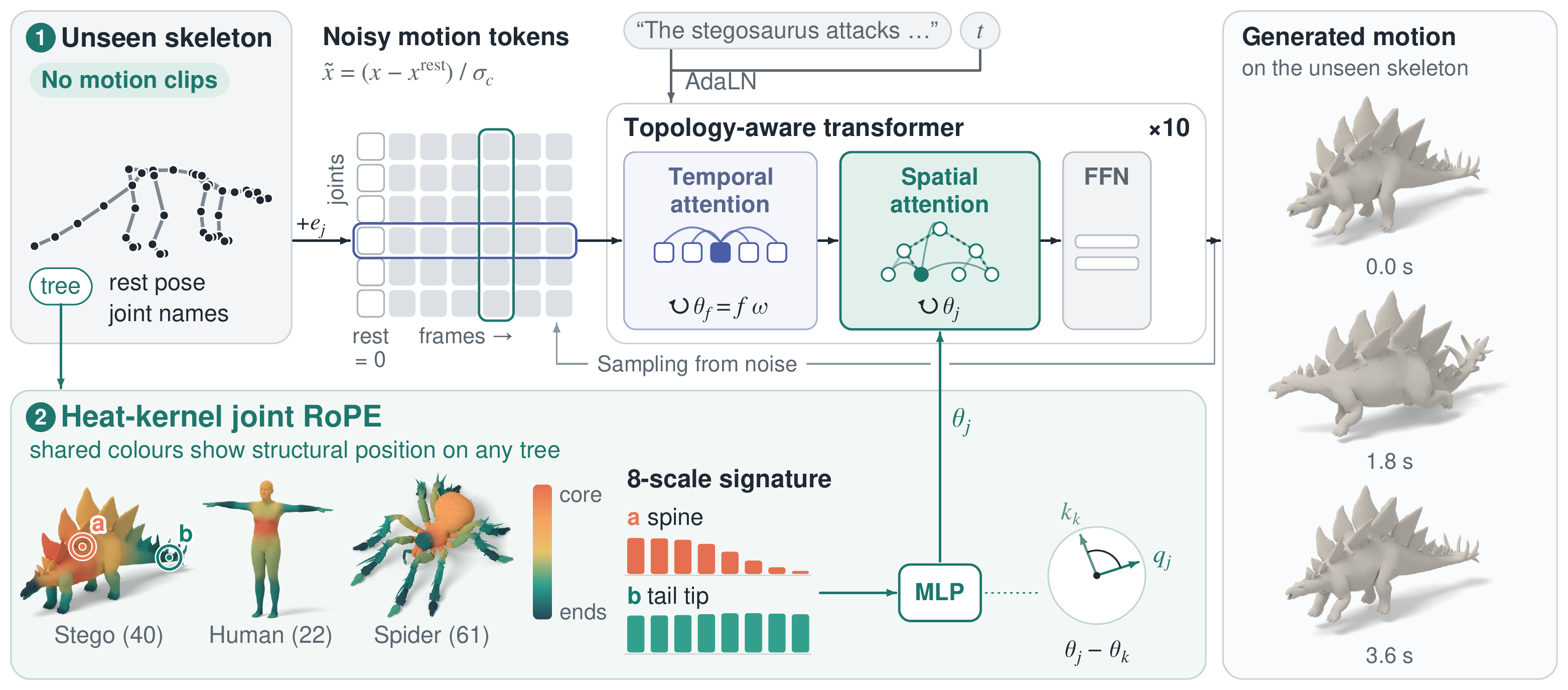}
\caption{\method. \ding{182}~No clips of an unseen skeleton are needed. Its tree, rest pose and joint names give
each joint a feature $e_j$ (an MLP of tree and rest-pose features plus the LLM2Vec embedding of the joint's description). The
motion tokens, one per joint and frame (position, rotation relative to the rest pose, velocity and contact; ground track and
heading on the root), are normalised by the rest pose and the skeleton's size (\cref{eq:restnorm}), so that the rest frame maps
to zero. \ding{183}~Each of ten blocks attends over time with queries and keys rotated by $\theta_f=f\omega$, and over joints with queries and keys rotated by HK-RoPE,
whose angle $\theta_j$ is an MLP of joint $j$'s heat-kernel signature at eight scales on the tree, together with a bias on
tree hops. Mesh colours show the first principal component of the log signature, on a colour scale shared by all
skeletons. The caption and the flow time $t$ condition the network through AdaLN. Sampling from noise yields the motion
shown on Stego, a Truebones test skeleton. Meshes are for display only; the human mesh is SMPL+H~\cite{romero2017mano}.}\label{fig:arch}
\end{figure*}

A rig $\mathcal{R}=(\mathcal{T},\Prest,\mathcal{N})$ consists of a kinematic tree $\mathcal{T}$ over $J$ joints, a rest pose
(global joint positions $\Prest\in\R^{J\times3}$ and rest orientations $\mathbf{R}^{\mathrm{rest}}_j$), and joint names
$\mathcal{N}$. Given $\mathcal{R}$ and a caption $c$, \method{} samples a motion $\x\in\R^{L\times J\times17}$ from
$p_\theta(\x\mid\mathcal{R},c)$ using a single transformer shared across all rigs. At generation time, no network input depends on a
rig's clips, so a rig unseen in training is handled exactly like a training rig. We apply this clip-free generation to
every benchmark skeleton, whether seen or unseen in training.

\subsection{Representation and skeleton inputs}\label{sec:method:rep}

\paragraph{Per-joint channels.} A representation shared by all rigs must meet two requirements: it must be computable
for any rigged animation from the kinematic tree, the rest pose and the motion alone, without a template skeleton,
retargeting or inverse kinematics; and each channel must have the same meaning on every rig. We therefore describe every
joint by the same 17 channels (\cref{tab:channels} in the appendix): (i)~its position, with horizontal coordinates
relative to the smoothed ground track of the root and absolute height, which decouples body pose from global displacement;
(ii)~its global rotation relative to its rest orientation, $\mathbf{R}_j(f)\,\mathbf{R}^{\mathrm{rest}\top}_j$ at frame $f$,
in the continuous 6D form~\cite{zhou2019continuity}, which equals the identity at the rest pose on every rig; (iii)~its
world velocity, used only as a supervision target; and (iv)~a binary contact flag. The root additionally carries the
ground track and the heading. Since velocities are never integrated, each frame can be decoded independently and without
accumulated drift, either from the positions or by forward kinematics (FK) of the rotations, through two independent
paths. The rotations and the root trajectory drive the rig, and all statistics in this paper are computed from them. The
same conversion places the four corpora of \bench in a common representation (\cref{sec:bench:data}).

\paragraph{Rest normalisation.} Rig sizes span several orders of magnitude, and a model without access to a rig's
clips cannot standardise its motion by their statistics. We centre every entry $\x_{fjc}$ (frame $f$, joint $j$, channel
$c$) on the rig's rest frame and scale it by the rig's size:
\begin{equation}
  \tilde{\x}_{fjc}=\frac{\x_{fjc}-\x^{\mathrm{rest}}_{jc}}{\sigma_c},
  \label{eq:restnorm}
\end{equation}
where $\x^{\mathrm{rest}}$ is the rest pose written in the 17 channels (rest positions, identity rotations, zero velocity and
contact, forward heading), and $\sigma_c=s/g_c$ on the position, velocity and ground-track channels, where $s$ is the diagonal of the rest pose's bounding box and $g_c$ is one of three global gains,
and $\sigma_c=1$ on all other channels. At generation, the validity mask of an unseen rig follows from its structure alone: every joint carries the
per-joint channels, and the root additionally carries the global ones.

\paragraph{Per-joint features.} Three features describe each joint and are added to all of its tokens
(\cref{sec:method:model}). (i)~A \emph{description}, a short phrase such as ``left thigh joint'', written from the joint's name or,
when the name is unrecognised or carries no anatomical meaning, from the tree's geometry and, where annotated, the body
plan; it is embedded once by a frozen LLM2Vec encoder~\cite{behnamghader2024llm2vec}. Descriptions make joint semantics
comparable across rigs: ``front left upper leg'' denotes the same concept on a horse and on a lizard, whereas raw names
such as \texttt{L\_Femur\_02} and \texttt{b\_hindleg\_l\_1} do not. (ii)~A \emph{structural vector} comprising the rest offset to the parent in
the parent's frame, the bone length, the depth, the path length to the root, the number of children and a leaf flag,
mapped through a zero-initialised MLP. (iii)~A \emph{world-frame rest
descriptor}: the world direction of the joint's bone at rest and its rest position relative to the root's ground point, expressed in units of the mean bone length and log-compressed. The parent-frame offset in (ii) depends on how the rig's author oriented each joint's
rest frame; the world-frame descriptor does not, and it is the feature that conveys the rest pose to the network.

\subsection{Factorised attention with HK-RoPE}\label{sec:method:model}

A motion is a grid of tokens over two axes of different kinds: the joints of a frame form a tree that varies from rig to rig, whereas the frames of a joint form a sequence. \method{} factorises attention along the two axes: each of its $N_b$ blocks
attends over time (one joint across its frames), then over joints (one frame across its joints), then applies an MLP.
Position is encoded on both axes in the same way: queries and keys are RMS-normalised per head~\cite{dehghani2023vit22b} and
rotated by angles of their token's position~\cite{su2024roformer}, so that position affects an attention logit only through the difference between the angles
of the two tokens.  The two encodings are designed jointly for skeleton graphs of arbitrary topology, and neither axis uses a learned table of
absolute positions.

\paragraph{Tokens and the rest frame.} Each (frame, joint) pair forms one token: its 17 normalised channels are projected to
width $D$. As in AnyTop~\cite{gat2025anytop}, the rest pose is prepended to the sequence as a separate frame: the input
$[\,\x^{\mathrm{rest}}\mid\x^{\mathrm{tgt}}_t\,]$ consists of one clean rest frame, which is never noised, followed by $L$
target frames that carry the flow interpolant and a learned marker; only the target frames are denoised. Under~\cref{eq:restnorm}, the rest frame is zero in every channel, so the target frames encode displacements from the
rest pose of the given skeleton; the shape of the skeleton itself reaches the network through the per-joint features of
\cref{sec:method:rep}.

\paragraph{Joint axis: HK-RoPE and tree bias.} The joint axis requires a positional coordinate that (i)~is defined
for every tree, (ii)~is invariant to the order in which the file lists the joints and to the eigenbasis returned by an
eigensolver, and (iii)~has the same dimension for any joint count, so that a single model applies to every skeleton.
HK-RoPE (heat-kernel RoPE) derives this coordinate from the heat kernel of the skeleton graph. Let $\mathbf{L}=\mathbf{I}-\mathbf{D}^{-1/2}\mathbf{A}\mathbf{D}^{-1/2}$
be the normalised Laplacian of the tree, with eigenpairs $(\lambda_k,\mathbf{v}_k)$, $0=\lambda_0<\lambda_1\le\dots\le\lambda_{J-1}$.
For eight diffusion times $\tau_i=2^{i-2}$, $i=0,\dots,7$,
\begin{equation}
  h_j(\tau)=\sum_{k=1}^{J-1}e^{-\tau\lambda_k}\,v_{kj}^2,\qquad
  \tilde h_j(\tau)=\frac{h_j(\tau)}{\tfrac1J\sum_{i}h_i(\tau)},
  \label{eq:hks}
\end{equation}
and an MLP maps $\tilde{\mathbf{h}}_j\in\R^8$ to the $d_h/2$ angles $\boldsymbol\theta_j$ of joint $j$. The spatial attention of
head $h$ between joints $i$ and $j$ of a frame is then
\begin{equation}
    \ell^{(h)}_{ij}=\frac{\mathbf{q}_i^{\top}\mathbf{R}(\boldsymbol\theta_i)^{\!\top}\mathbf{R}(\boldsymbol\theta_j)\,\mathbf{k}_j}{\sqrt{d_h}}+b^{(h)}_{ij}
  =\frac{\mathbf{q}_i^{\top}\mathbf{R}(\boldsymbol\theta_j-\boldsymbol\theta_i)\,\mathbf{k}_j}{\sqrt{d_h}}+b^{(h)}_{ij},
  \label{eq:jointrope}
\end{equation}
where $\mathbf{R}(\boldsymbol\theta)$ rotates the $d_h/2$ coordinate pairs of a query or key by the angles $\boldsymbol\theta$, so the
joint coordinate enters only through $\boldsymbol\theta_j-\boldsymbol\theta_i$. The bias $b^{(h)}_{ij}$, in the spirit of the
graph-distance bias of AnyTop~\cite{gat2025anytop}, is $b^{(h)}_{ij}=-\min(g_{ij},8)+u^{(h)}_{a_{ij}}+d^{(h)}_{s_{ij}}$; it
depends on the hop distance $g_{ij}$ and, through zero-initialised tables $u$ and $d$, on the numbers of hops $a_{ij}$ up
and $s_{ij}$ down to the lowest common ancestor, so that a head can distinguish a parent from its child and a limb from its
counterpart on the other side of the body. $h_j(\tau)$ is the heat-kernel signature~\cite{sun2009hks,canturk2024gpse} with the zero-eigenvalue mode removed, and
it satisfies all three requirements. It is defined on every tree and has eight values per joint regardless of $J$: all
$J-1$ non-trivial modes contribute, so no truncation or padding is needed. It does not depend on the
eigenbasis: summed over an eigenspace, $v_{kj}^2$ is the diagonal of the eigenspace's projector, the same for every basis and
every sign. Finally, relabelling the joints merely permutes its rows; since every other input is attached to a joint rather than to a
slot, the motion predicted for each joint does not depend on the order in which the file lists the joints (proof in the
appendix). In contrast, spectral rotary encodings of graphs derive their angles from Laplacian
eigenvectors~\cite{reid2026wire}; UniMate's spectral joint encoding~\cite{mou2026unimate} resolves their sign ambiguity with
a SignNet~\cite{lim2023signnet}, but neither the basis ambiguity within a repeated eigenvalue nor the one introduced when
truncation splits an eigenspace. This case is common: 3,210 of our 5,127 training skeletons (63\%) have a repeated eigenvalue among the first eight non-trivial modes or
between the eighth and the ninth.

\paragraph{Frame axis.} The temporal attention rotates queries and keys in the same way~\cite{su2024roformer}, by
$\boldsymbol\theta_f=f\,\boldsymbol\omega$ with $\omega_m=700^{-2m/d_h}$, $m=0,\dots,d_h/2-1$, for frame index $f$ (the rest frame has $f=0$); the base of 700 is chosen for a 241-frame window.

\paragraph{Text and time.} The conditioning vector $\mathbf{c}=\mathrm{MLP}(t)+\mathrm{MLP}(\mathbf{e}_c)$, where $t$ is the flow time and
$\mathbf{e}_c$ the pooled LLM2Vec embedding of the caption, produces the AdaLN shift, scale and gate of every
sub-layer~\cite{peebles2023dit}; the text thus conditions the network through modulation only. The caption is dropped with probability 0.1 for
classifier-free guidance~\cite{ho2022cfg}; the rest frame is never dropped.

\subsection{Training and sampling}\label{sec:method:train}

\paragraph{Objective.} We train on the straight path $\x_t=(1-t)\bm{\epsilon}+t\,\x$,
$\bm{\epsilon}\sim\mathcal{N}(0,\mathbf{I})$, $t\sim\mathcal{U}(0,1)$~\cite{lipman2023flow,liu2023rectified}; the network
predicts the clean sample $\hat\x$ and is trained with a per-entry Huber loss, weighted by the velocity-space factor
$\max(1-t,\sigma_{\min})^{-2}$ of JiT~\cite{li2025jit} and normalised to unit mean over $t$. A 142-joint rig has 28 times as many body rows as
a six-joint rig but the same single root row, so a plain mean over entries would let the body rotations of large rigs
overwhelm the root trajectory and contacts of small ones. We partition the channels into nine groups $g$ (position, rotation and velocity of the root and
of the body; contact; ground track; heading) and use
\begin{equation}
  \mathcal{L}=\sum_{g}\gamma_g\sqrt{N_g/N}\;\bar\ell_g,
  \label{eq:grouped}
\end{equation}
where $\bar\ell_g$ is the masked mean over the $N_g$ valid entries of group $g$ in the batch and $N=\sum_gN_g$, so that a
group's share of the gradient does not grow with the number of entries it spans. The weights $\gamma_g$ are computed from measured
group energies so that each group receives a fixed target share of the objective (appendix). Auxiliary FK-consistency, world-velocity, foot-lock and acceleration terms are
also added.

\paragraph{Skeleton and convention augmentation.} With probability 0.8, a training sample undergoes one
kinematics-preserving operation: a joint is inserted on a bone, a few leaves are removed, single-child joints of a chain
are merged, or every bone is rescaled by up to $\pm10\%$; the sample is then re-encoded by FK to remain self-consistent.
Independently, with probability 0.5, every joint's rest orientation is rotated randomly by up to $30^\circ$, as if the rig
had been authored with another rest pose: the world motion is unchanged, and only the rest-relative rotations that the
network must predict change. The changed rest pose reaches the network through the world-frame rest descriptor. Rigs are sampled with
probability proportional to the square root of their clip count, so that corpora of single-clip rigs and of rigs with
hundreds of clips both contribute.

\paragraph{Sampling.} Sampling integrates the straight path with $S$ Euler steps, $\x\leftarrow\x+(\hat\x-\x)/(S-i)$
at step $i$; invalid entries are kept at zero, and classifier-free guidance is applied to the caption only:
$\hat\x=\hat\x^{\varnothing}+w\,(\hat\x^{c}-\hat\x^{\varnothing})$. The output is used as generated, without grounding,
foot locking or inverse kinematics.

\section{The \bench corpus and benchmark}\label{sec:bench}

\subsection{Four corpora under one codec}\label{sec:bench:data}

\bench unifies four corpora under one codec (counts refer to training clips and rigs after the benchmark split is
removed):
the Objaverse-XL~\cite{deitke2023objaversexl} subset of UniML3D~\cite{mou2026unimate} at a fixed commit (5,023 clips, 4,277
rigs, 6--60 joints), with one caption per clip written by a vision--language model; the AniMo4D animal
library~\cite{wang2025animo} (76,882 clips, 307 rigs) with its captions, together with HumanML3D~\cite{guo2022generating}
(26,844 clips on one 22-joint skeleton); further animations and species from the same game library as AniMo4D (29,645 clips,
485 rigs), captioned by rule from the animation names; and the Truebones zoo~\cite{truebones} (708 clips, 57 rigs, up to 142
joints). Training thus covers 139,102 clips (144 hours at 30\,fps) on 5,127 rigs and 1,364 distinct trees, with 6 to 142
joints. All clips are encoded at 30\,fps by the same codec, grounded at
their lowest point, labelled with the same contact rule and passed through a data-quality check (appendix), so that every
statistic has the same meaning across corpora.  Since the source assets retain their original licences, the corpus will be released as per-clip source identifiers,
splits and annotations, together with scripts that rebuild it from the originals and the metric code.

\begin{table*}[t]
\centering\small
\caption{\bench buckets. A bucket holds out \emph{units}---clips, rigs, classes of isomorphic trees or species---and every clip
of a held-out unit is removed from training and validation. TED: median tree edit distance from a clip's tree to the nearest
training tree. Reused motion: share of clips whose source animation appears in training (detected by motion fingerprints for UniML3D and by matching joint rotations for the game species).}\label{tab:buckets}
\setlength{\tabcolsep}{3.6pt}
\begin{tabular}{@{}lllrrrrrr@{}}
\toprule
Bucket & Source & Unseen at test time & Clips dev / test & Rigs & Trees & Joints & TED & Reused motion \\
\midrule
ID-clip            & UniML3D   & clips of training rigs                     & 70 / 172   & 135 & 79  & 10--60  & 0  & 26\% \\
ID-tree            & UniML3D   & rigs whose tree is seen                    & 255 / 508  & 759 & 140 & 6--60   & 0  & 63\% \\
\midrule
OOD-tree           & UniML3D   & tree classes (bipeds, quadrupeds)          & 213 / 320  & 333 & 266 & 11--60  & 2  & 15\% \\
OOD-plan           & UniML3D   & tree classes of rare body plans            & 40 / 62    & 70  & 63  & 5--60   & 5  & 2\% \\
OOD-J$>$60 (Mixamo)& UniML3D   & rigs of one 65-joint humanoid tree         & 155 / 350  & 250 & 1   & 65      & 5  & 51\% \\
OOD-J$>$60 (other) & UniML3D   & rigs above the corpus's joint range        & 166 / 512  & 464 & 184 & 61--117 & 7  & 33\% \\
OOD-J$<$5          & UniML3D   & rigs below that range                      & 1 / 1      & 2   & 2   & 4       & 2  & 0\% \\
OOD-species        & game      & four species, three rigs each              & 630 / 635  & 12  & 8   & 41--66  & 5  & 18\% \\
OOD-new-species    & game      & rigs of ten untrained species              & 1 / 47     & 10  & 10  & 28--78  & 11 & -- \\
OOD-rig            & Truebones & eight quadruped rigs                       & 32 / 62    & 8   & 8   & 31--49  & 4  & -- \\
\midrule
Total              &           &                                            & 1,563 / 2,669 & 2,043 & 736 & 4--117 &  &  \\
\bottomrule
\end{tabular}
\end{table*}

\subsection{Held-out buckets}\label{sec:bench:split}

Each \bench bucket holds out a set of units (\cref{tab:buckets}). Two in-distribution buckets serve as references:
\emph{ID-clip} holds out clips of training rigs, and \emph{ID-tree} holds out whole rigs whose tree is isomorphic
to a training rig's, so that only proportions, rest pose and names change. Each out-of-distribution bucket shifts along one axis. \emph{OOD-tree} holds out a quarter of the rarely populated tree classes of bipeds and quadrupeds, and
\emph{OOD-plan} 40\% of the tree classes of rare body plans (insects, birds, sea animals, snakes); every rig of a held-out class is removed from training. The \emph{OOD-J} buckets contain the UniML3D rigs whose joint count lies outside that corpus's training range:
61--117 joints (250 of which share Mixamo's 65-joint humanoid tree) and 4 joints. \emph{OOD-species} holds out four
species of the game library---markhor and greater rhea (dev), capybara and emu (test)---with their male, female and juvenile
rigs. \emph{OOD-new-species} contains ten rigs of species absent from training (frogs, a newt, an axolotl, beetles, a cockroach, a leaf insect, a fruit bat, a terrapin), retained after a motion-quality check.
\emph{OOD-rig} contains eight Truebones quadrupeds (two bears, cat, goat, hound, lynx, stegosaurus, turtle) with unseen trees.

Except in OOD-species, held-out units are assigned by hashing, one third to dev and two thirds to test; model selection
and hyper-parameter choices for \method{} use the dev split only.  Every benchmark clip is annotated with the distance from its tree to the nearest training tree, its joint count
and body plan, and whether its source motion was reused on a training skeleton, so that results can be stratified by any
of these annotations.

\subsection{Protocol and metrics}\label{sec:bench:metrics}

\paragraph{Generation.} \method{} animates every benchmark rig, including those of ID-clip, from its skeleton file: the tree, the rest pose,
the joint names and the up and forward axes. For each clip, five samples are generated with different seeds at the length of the real clip (capped at 240 frames),
using the guidance scale with the lowest \kidk{} on dev (appendix), and scored without post-processing.

\paragraph{Motion quality.} We measure Fréchet distances on kinematic features, computed on the animation that drives
the rig, \ie the FK of the generated rotations together with the root trajectory:
\begin{itemize}
\item \fidk: Fréchet distance between the generated and the real clips of a group, computed on the public kinetic
  features of Bailando~\cite{siyao2022bailando} and Motion2Motion~\cite{chen2025motion2motion}, which UniMate's FID also
  uses~\cite{mou2026unimate}: per joint, the mean horizontal and vertical kinetic energy and the mean acceleration over the
  clip, with positions in units of the rig scale. Pooling them over joints by mean and standard deviation yields six
  features for every skeleton; both sets are standardised by the real clips, and each group must contain at least 50 real
  clips.
\item \fidd: the same distance on the 24 kinematic descriptors of our scorer (below), which also cover pose, contact,
  rhythm and path.
\end{itemize}
Per bucket, our scorer (appendix) reports \kidk, a kernel distance~\cite{binkowski2018kid} on the same 24
kinematic descriptors per clip (speeds, accelerations, contact, turning, pose spread, path), standardised by the training
clips, which remains well defined on small groups and is reported together with its real--real value, \ie the KID
between two halves of the real clips; Slip, the share of grounded joint-frames that slide, computed over all joints using
the codec's contact rule; Pen., the share of frames whose lowest joint is more than $0.005\,s$ below the ground; Float,
the share of clips (excluding those with airborne captions) that hover higher than 95\% of the training clips of their
body plan; and Speed, the joint speed relative to the real clips. All statistics are averaged over seeds and weight rigs
equally, except \kidk, which compares clip distributions.

\begin{figure*}[t]
\centering
\includegraphics[width=0.91\textwidth]{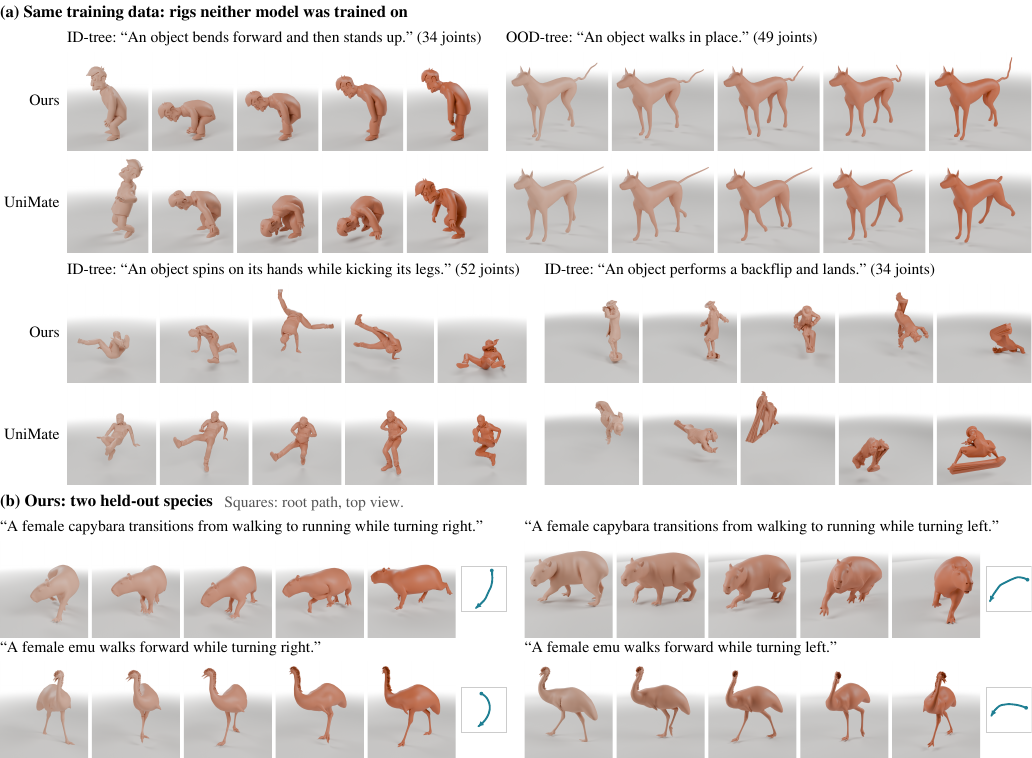}
\caption{Samples on unseen test skeletons, each drawn on the rig's own mesh: five frames at equal times, light to dark.
\textbf{(a)}~The two models of \cref{tab:main} (UniMate tuned on \fidk) on four rigs that neither was trained on,
\method{} in the upper row, over at most 60 frames. \textbf{(b)}~The main model on two held-out game species, a capybara and an emu on their production meshes; the camera
follows the animal, and squares show its root path from above. Rendering details, further examples and mesh
credits are given in \cref{app:qual}.}\label{fig:qual}
\end{figure*}

\paragraph{Text alignment.} We measure text alignment by text-to-motion retrieval with a skeleton-agnostic evaluator
trained on our training split (appendix), within the rigs of the held-out game species (\nTaAllClips{} test clips with
curated captions, pools of 8, chance level 0.125). \method{} reaches R@1 \nTaMainClipFullAllR, compared with
\nTaRealFullAllR{} for the real motions and \nTaMainClipFullAllShuf{} with shuffled captions. \Cref{fig:qual} shows
qualitative examples.

\begin{table*}[t]
\centering\footnotesize
\caption{Comparison of \method{} and UniMate, both trained on the same 5,023 UniML3D clips (test clips of the four UniML3D buckets, first
60 frames, five seeds; additional metrics: appendix),
with UniMate's guidance tuned on dev by \fidk{} (left) or by \fidd{} (right). Bold: the lower value within each half, where the paired 95\%
confidence interval over rigs excludes zero. R--R: \fidd{} between two halves of the real clips. Pen.: share of frames below the ground (0 for real
clips).}\label{tab:main}
\setlength{\tabcolsep}{4pt}
\begin{tabular}{@{}lrc|cccccc|cccccc@{}}
\toprule
& & & \multicolumn{6}{c|}{UniMate tuned on \fidk} & \multicolumn{6}{c}{UniMate tuned on \fidd} \\
& & & \multicolumn{2}{c}{\fidd$\downarrow$} & \multicolumn{2}{c}{\fidk$\downarrow$} & \multicolumn{2}{c|}{Pen.$\downarrow$}
  & \multicolumn{2}{c}{\fidd$\downarrow$} & \multicolumn{2}{c}{\fidk$\downarrow$} & \multicolumn{2}{c}{Pen.$\downarrow$} \\
Test clips & Clips & R--R & Ours & Uni. & Ours & Uni. & Ours & Uni. & Ours & Uni. & Ours & Uni. & Ours & Uni. \\
\midrule
All                    & \nTwoAllClips & \nTwoAllDescFloor & \nTwoAllDescOursB & \nTwoAllDescUniB & \nTwoAllKinOursB & \nTwoAllKinUniB & \nTwoAllPenOursB & \nTwoAllPenUniB & \nThrAllDescOursB & \nThrAllDescUniB & \nThrAllKinOursB & \nThrAllKinUniB & \nThrAllPenOursB & \nThrAllPenUniB \\
ID-clip                & \nTwoClipClips & \nTwoClipDescFloor & \nTwoClipDescOursB & \nTwoClipDescUniB & \nTwoClipKinOursB & \nTwoClipKinUniB & \nTwoClipPenOursB & \nTwoClipPenUniB & \nThrClipDescOursB & \nThrClipDescUniB & \nThrClipKinOursB & \nThrClipKinUniB & \nThrClipPenOursB & \nThrClipPenUniB \\
No training clip       & \nTwoNoTrClips & \nTwoNoTrDescFloor & \nTwoNoTrDescOursB & \nTwoNoTrDescUniB & \nTwoNoTrKinOursB & \nTwoNoTrKinUniB & \nTwoNoTrPenOursB & \nTwoNoTrPenUniB & \nThrNoTrDescOursB & \nThrNoTrDescUniB & \nThrNoTrKinOursB & \nThrNoTrKinUniB & \nThrNoTrPenOursB & \nThrNoTrPenUniB \\
\quad novel motion     & \nTwoNovelClips & \nTwoNovelDescFloor & \nTwoNovelDescOursB & \nTwoNovelDescUniB & \nTwoNovelKinOursB & \nTwoNovelKinUniB & \nTwoNovelPenOursB & \nTwoNovelPenUniB & \nThrNovelDescOursB & \nThrNovelDescUniB & \nThrNovelKinOursB & \nThrNovelKinUniB & \nThrNovelPenOursB & \nThrNovelPenUniB \\
\quad reused motion    & \nTwoReusedClips & \nTwoReusedDescFloor & \nTwoReusedDescOursB & \nTwoReusedDescUniB & \nTwoReusedKinOursB & \nTwoReusedKinUniB & \nTwoReusedPenOursB & \nTwoReusedPenUniB & \nThrReusedDescOursB & \nThrReusedDescUniB & \nThrReusedKinOursB & \nThrReusedKinUniB & \nThrReusedPenOursB & \nThrReusedPenUniB \\
\quad ID-tree          & \nTwoSTreeClips & \nTwoSTreeDescFloor & \nTwoSTreeDescOursB & \nTwoSTreeDescUniB & \nTwoSTreeKinOursB & \nTwoSTreeKinUniB & \nTwoSTreePenOursB & \nTwoSTreePenUniB & \nThrSTreeDescOursB & \nThrSTreeDescUniB & \nThrSTreeKinOursB & \nThrSTreeKinUniB & \nThrSTreePenOursB & \nThrSTreePenUniB \\
\quad OOD-tree         & \nTwoTreeClips & \nTwoTreeDescFloor & \nTwoTreeDescOursB & \nTwoTreeDescUniB & \nTwoTreeKinOursB & \nTwoTreeKinUniB & \nTwoTreePenOursB & \nTwoTreePenUniB & \nThrTreeDescOursB & \nThrTreeDescUniB & \nThrTreeKinOursB & \nThrTreeKinUniB & \nThrTreePenOursB & \nThrTreePenUniB \\
\quad OOD-plan         & \nTwoPlanClips & -- & \nTwoPlanDescOursB & \nTwoPlanDescUniB & \nTwoPlanKinOursB & \nTwoPlanKinUniB & \nTwoPlanPenOursB & \nTwoPlanPenUniB & \nThrPlanDescOursB & \nThrPlanDescUniB & \nThrPlanKinOursB & \nThrPlanKinUniB & \nThrPlanPenOursB & \nThrPlanPenUniB \\
\bottomrule
\end{tabular}
\end{table*}

\begin{table*}[t]
\centering\footnotesize
\caption{Results of \method{} on every test bucket at full length (up to 240 frames, five seeds per clip). Real-clip
values are in parentheses ($\rightarrow$: closer to the real value is better), and \kidk{} is shown with its real--real value. $^\ast$Fewer than 20 rigs. The single OOD-J$<$5 test clip is included only in the last
row.}\label{tab:perbucket}
\setlength{\tabcolsep}{6pt}
\begin{tabular}{@{}lrccccc@{}}
\toprule
Bucket & Rigs / clips & \kidk$\downarrow$ (real--real) & Slip$\rightarrow$ (real) & Pen.$\downarrow$ & Float$\rightarrow$ (real) & Speed$\rightarrow$ \\
\midrule
ID-clip                 & \nPbIDclipRigs{} / \nPbIDclipClips   & \nPbIDclipKID{} (\nPbIDclipFloor)   & \nPbIDclipSlip{} (\nPbIDclipSlipReal)   & \nPbIDclipPen   & \nPbIDclipFloat{} (\nPbIDclipFloatReal)   & \nPbIDclipSpeed \\
ID-tree                 & \nPbIDtreeRigs{} / \nPbIDtreeClips   & \nPbIDtreeKID{} (\nPbIDtreeFloor)   & \nPbIDtreeSlip{} (\nPbIDtreeSlipReal)   & \nPbIDtreePen   & \nPbIDtreeFloat{} (\nPbIDtreeFloatReal)   & \nPbIDtreeSpeed \\
\midrule
OOD-tree                & \nPbTreeRigs{} / \nPbTreeClips       & \nPbTreeKID{} (\nPbTreeFloor)       & \nPbTreeSlip{} (\nPbTreeSlipReal)       & \nPbTreePen     & \nPbTreeFloat{} (\nPbTreeFloatReal)       & \nPbTreeSpeed \\
OOD-plan                & \nPbPlanRigs{} / \nPbPlanClips       & \nPbPlanKID{} (\nPbPlanFloor)       & \nPbPlanSlip{} (\nPbPlanSlipReal)       & \nPbPlanPen     & \nPbPlanFloat{} (\nPbPlanFloatReal)       & \nPbPlanSpeed \\
OOD-J$>$60 (Mixamo)     & \nPbMixamoRigs{} / \nPbMixamoClips   & \nPbMixamoKID{} (\nPbMixamoFloor)   & \nPbMixamoSlip{} (\nPbMixamoSlipReal)   & \nPbMixamoPen   & \nPbMixamoFloat{} (\nPbMixamoFloatReal)   & \nPbMixamoSpeed \\
OOD-J$>$60 (other)      & \nPbJotherRigs{} / \nPbJotherClips   & \nPbJotherKID{} (\nPbJotherFloor)   & \nPbJotherSlip{} (\nPbJotherSlipReal)   & \nPbJotherPen   & \nPbJotherFloat{} (\nPbJotherFloatReal)   & \nPbJotherSpeed \\
OOD-species$^\ast$      & \nPbSpeciesRigs{} / \nPbSpeciesClips & \nPbSpeciesKID{} (\nPbSpeciesFloor) & \nPbSpeciesSlip{} (\nPbSpeciesSlipReal) & \nPbSpeciesPen  & \nPbSpeciesFloat{} (\nPbSpeciesFloatReal) & \nPbSpeciesSpeed \\
OOD-new-species$^\ast$  & \nPbNewSpRigs{} / \nPbNewSpClips     & \nPbNewSpKID{} (\nPbNewSpFloor)     & \nPbNewSpSlip{} (\nPbNewSpSlipReal)     & \nPbNewSpPen    & \nPbNewSpFloat{} (\nPbNewSpFloatReal)     & \nPbNewSpSpeed \\
OOD-rig$^\ast$          & \nPbRigRigs{} / \nPbRigClips         & \nPbRigKID{} (\nPbRigFloor)         & \nPbRigSlip{} (\nPbRigSlipReal)         & \nPbRigPen      & \nPbRigFloat{} (\nPbRigFloatReal)         & \nPbRigSpeed \\
\midrule
All buckets             & \nPbAllRigs{} / \nPbAllClips         & \nPbAllKID{} (\nPbAllFloor)         & \nPbAllSlip{} (\nPbAllSlipReal)         & \nPbAllPen      & \nPbAllFloat{} (\nPbAllFloatReal)         & \nPbAllSpeed \\
\bottomrule
\end{tabular}
\end{table*}

\section{Experiments}\label{sec:exp}

\subsection{Setup}\label{sec:exp:setup}

\paragraph{Model.} \method{} consists of 10 blocks of width 512 with 8 attention heads (73.3M
parameters). Each ablation is trained for half the steps of the main model and compared with a full-model control trained
for the same number of steps; all arms share the same sampler settings. Further training details are given in the
appendix.

\paragraph{Baseline.} UniMate~\cite{mou2026unimate}, the most closely related method, animates an unseen skeleton zero-shot from its rest pose,
topology and joint names, and generates 60 frames. We train both methods on the same 5,023 clips of
the UniML3D training split, UniMate with its official code and configuration, and \method{} as described in
\cref{sec:method}. Each method uses its own sampler, with \method{} generating at the length of the real clip; guidance scales are
selected on dev, for \method{} by the lowest \fidk{} and for UniMate by the lowest \fidk{} and, separately, the lowest
\fidd{} (appendix). A comparison with released weights, including those of AnyTop~\cite{gat2025anytop}, is given in the appendix.

\subsection{Skeletons without training clips}\label{sec:exp:main}

\Cref{tab:main} compares the two methods trained on the same clips. UniMate's guidance trades one distance for the
other: its dev optimum is \nTwoUniCfg{} under \fidk{} and \nThrUniCfg{} under \fidd, which on all test clips give \fidk{}
\nTwoAllKinUni{} with \fidd{} \nTwoAllDescUni, and \fidd{} \nThrAllDescUni{} with \fidk{} \nThrAllKinUni. \method{} has the
same dev optimum under both distances (guidance \nTwoOursCfg; appendix), at which it obtains \fidk{} \nTwoAllKinOurs{}
and \fidd{} \nTwoAllDescOurs. On rigs unseen in training whose tree is isomorphic to a training rig (ID-tree) and on
reused motion, \method{} is lower on both distances under both tunings (ID-tree \fidd{} \nTwoSTreeDescOurs{} against
\nTwoSTreeDescUni{} and \nThrSTreeDescUni), and under either tuning fewer of its frames fall below the ground
(\nTwoAllPenOurs{} against \nTwoAllPenUni{} and \nThrAllPenUni).
With released weights, on the two held-out species whose rigs none of the three models was trained on, \method{}
attains \fidd{} \nSpOursFID, compared with \nSpUniFID{} for UniMate and \nSpAnyFID{} for AnyTop (\cref{app:released}).
\Cref{fig:qual} shows qualitative samples, and \cref{app:qual} provides further examples. \Cref{tab:perbucket}
reports \method{} on every test bucket at full length, where \kidk{} is about equal to its real--real value on the
held-out game species (\nPbSpeciesKID{} against \nPbSpeciesFloor) and below it on Mixamo's 65-joint tree
(\nPbMixamoKID{} against \nPbMixamoFloor). 

\subsection{Ablations}\label{sec:exp:ablation}

\begin{table}[t]
\centering\footnotesize
\caption{Ablations, each trained for half the steps of the main model and evaluated on the OOD buckets targeted by the
changed components:
\kidk{} on unseen topologies (OOD-tree), rare body plans (OOD-plan), new species and Truebones rigs (OOD-rig), five seeds per
clip, up to 240 frames. Rest: median change of the output, in mean bone lengths, when a test rig's rest convention is rotated
(\cref{app:abl}). Real: \kidk{} between two halves of the real clips. Confidence half-widths are given in \cref{app:abl}.}\label{tab:ablation}
\setlength{\tabcolsep}{3.5pt}
\begin{tabular}{@{}lccccc@{}}
\toprule
Arm & \shortstack{OOD-\\tree} & \shortstack{OOD-\\plan} & \shortstack{New\\species} & \shortstack{OOD-\\rig} & Rest$\downarrow$ \\
\midrule
Real                   & \nAbTreeFloor & \nAbPlanFloor & \nAbNewSpFloor & \nAbRigFloor & -- \\
Full model             & \textbf{\nAbCtlTreeKID} & \textbf{\nAbCtlPlanKID} & \textbf{\nAbCtlNewSpKID} & \textbf{\nAbCtlRigKID} & \textbf{\nRcCtl} \\
\midrule
Learned frame table    & \nAbTroTreeKID & \nAbTroPlanKID & \nAbTroNewSpKID & \nAbTroRigKID & -- \\
Learned joint table    & \nAbJtabTreeKID & \nAbJtabPlanKID & \nAbJtabNewSpKID & \nAbJtabRigKID & -- \\
Eigenvectors + SignNet & \nAbEigTreeKID & \nAbEigPlanKID & \nAbEigNewSpKID & \nAbEigRigKID & -- \\
\midrule
$-$ augmentation       & \nAbAugTreeKID & \nAbAugPlanKID & \nAbAugNewSpKID & \nAbAugRigKID & \nRcAug \\
\bottomrule
\end{tabular}
\end{table}

All arms share the representation of \cref{sec:method:rep}, which is required to animate a rig from its file alone.
Each arm in \cref{tab:ablation} changes one component: the frame or the joint rotary encoding is replaced by a learned
position table, the heat-kernel coordinate of HK-RoPE is replaced by Laplacian eigenvectors with a
SignNet~\cite{lim2023signnet,mou2026unimate}, or the augmentation (of skeletons and of rest conventions) is removed. Since
these components target skeletons unlike those seen in training, we evaluate them on the buckets that test this: unseen
topologies, rare body plans, new species and Truebones rigs. On each of these buckets, the full model achieves the lowest
\kidk{} among all arms (on unseen topologies, \nAbCtlTreeKID{} compared with \nAbTroTreeKID{} with a frame table,
\nAbJtabTreeKID{} with a joint table, \nAbEigTreeKID{} with eigenvectors and \nAbAugTreeKID{} without augmentation), and
changing a test rig's rest convention alters its output only half as much as without augmentation (\nRcCtl{} against
\nRcAug{} mean bone lengths).

\section{Conclusion}\label{sec:conclusion}

We presented \method, a text-to-motion system that animates a skeleton unseen in training from its skeleton file alone. It
normalises motion by the rest pose and the skeleton's size, encodes the rest pose in the world frame under randomised
conventions, and encodes attention with HK-RoPE, which is defined for skeleton graphs of any topology. On \bench, which
measures generalisation along five axes of skeleton shift, a single model animates every skeleton from its file, and on
held-out species its motion is about as close to the real clips as two halves of the real clips are to each other. 

{\small
\bibliographystyle{plainnat}
\bibliography{main}}
\clearpage
\appendix
\renewcommand{\thesection}{\Alph{section}}
\raggedbottom

\section{Representation and codec}\label{app:rep}

\begin{table}[H]
\centering\footnotesize
\caption{Channels per joint and frame. Channels 13--16 are used on the root row only. Positions, velocities and the ground track
are in the skeleton's units; $s$ is the diagonal of the rest pose's bounding box.}\label{tab:channels}
\setlength{\tabcolsep}{3pt}
\begin{tabular}{@{}lll@{}}
\toprule
Slice & Content & Rows \\
\midrule
0:3   & position: XZ from the smoothed root track, Y absolute & all \\
3:9   & rotation $\mathbf{R}_j\mathbf{R}^{\mathrm{rest}\top}_j$, continuous 6D & all \\
9:12  & world velocity (supervision only) & all \\
12    & contact: height $\le0.05\,s$, speed $\le0.25\,s$/s & all \\
13:15 & smoothed root track XZ & root \\
15:17 & heading $(\cos\phi,\sin\phi)$ & root \\
\bottomrule
\end{tabular}
\end{table}

\paragraph{Codec.} A clip is given to the codec as a kinematic tree, a rest pose, and per-frame local rotations and root
translations. It is resampled
to 30\,fps (spherical interpolation of the local rotations, linear interpolation of the root), its global rotations and positions are obtained by FK, and it is shifted vertically so that its lowest joint over the whole clip touches $y{=}0$. The root's
horizontal track is smoothed by a fourth-order zero-phase Butterworth filter at 1\,Hz (a least-squares line for clips shorter
than 90 frames) and stored from the clip's first frame; positions store their horizontal coordinates relative to this
track and their absolute height. Rotations are
the global rotation composed with the inverse rest orientation, stored as the first two columns of the rotation matrix. The heading is the ground-plane projection of the forward axis of a designated carrier joint,
marked invalid where this projection vanishes.
The frame is right-handed with $+Y$ up and the rest pose facing $+Z$. A data-quality check verifies grounding, contact,
root track, heading and the agreement of the positions with the FK of the rotations on every clip.

\paragraph{Rest normalisation and validity.} The rest frame $\x^{\mathrm{rest}}$ of \cref{eq:restnorm} holds the rest
positions relative to the root's ground point, the identity rotation $[1,0,0,0,1,0]$, zero velocity, contact and track, and
the heading $(1,0)$; the gains $g=(3.868,\,2.944,\,3.321)$ for position, velocity and ground track are three constants, fixed once on a
calibration set of 148 clips and shared by all rigs. During training, a (joint, channel) pair that is exactly constant for a rig (for example, the contact flag of a joint that never touches the ground) is excluded from the loss and the input.

\section{Model details}\label{app:model}

\paragraph{Per-joint features.} The structural vector holds the unit rest offset to the parent in the parent's frame,
the bone length as $\log(1+\ell_j/\bar\ell)$, the depth divided by 16, the path length to the root as
$\log(1+p_j/\bar\ell)/4$, the number of children clipped at 4 and divided by 4, and a leaf flag, where $\bar\ell$ is the mean
bone length. The world-frame rest descriptor holds
$\mathbf{u}_j=(\Prest_j-\Prest_{\pi(j)})/\lVert\Prest_j-\Prest_{\pi(j)}\rVert$ (zero for the root) and
$\mathbf{c}_j=(\Prest_j-[P^{\mathrm{rest}}_{0,x},0,P^{\mathrm{rest}}_{0,z}])/\bar\ell$, compressed radially as
$\mathbf{c}\mapsto\mathbf{c}\log(1+\lVert\mathbf{c}\rVert)/\lVert\mathbf{c}\rVert$; it is fed through a bias-free linear map whose output is added to the first
layer of the structural MLP, whose last layer is zero-initialised. Descriptions are embedded by LLM2Vec
(4096-d) and projected to $D$.

\paragraph{Joint descriptions.} Every joint receives a name-based description (``Left Thigh joint.'') or a geometric description. On UniML3D, the name is the cleaned name provided by the corpus.
Built on 90\% of the UniML3D rigs, the lexicon described next reproduces the corpus's own description for \nLexExact\% of the name-described joints of the other 10\% (\nLexRigs{} rigs). All other rigs are described as a new rig is at deployment: a lexicon from raw to cleaned names over all UniML3D skeletons gives the name (a segment name of the game library first becomes the part it names, \texttt{frontLegUpr} the upper arm), and the side is taken from the rest pose ($+X$ is left beyond $0.02\,s$), even when the name indicates otherwise. A joint whose name has no entry or carries no anatomy
(``Bone'', ``Root'') is described from the rest pose and the tree instead: its side, the unbranched chain it lies on, its
place along it and whether the chain ends in a tip or a branch, its rank among sibling chains from front to back, its height
and depth relative to the root, and the rig's body plan (``the first joint of a six-bone limb on the left side of a
quadrupedal skeleton, below and behind the root''). The body plan is the corpus's annotation for benchmark rigs and is supplied by the user for a new rig, and otherwise defaults to ``a skeleton''.

\paragraph{Animating a new rig.} The model reads the tree, the rest pose and the joint names, with the up and forward axes of
the file, which rotate the rig into the frame of \cref{app:rep} ($+Y$ up, the rest pose facing $+Z$). 
The heading carrier joint is used only to encode real clips as training targets; generation does not use it.

\paragraph{Attention.} The tree bias clips the hop distance at 8 and the hop counts to the lowest common ancestor at 15;
padded joints receive a bias of $-10^4$. Queries and keys are RMS-normalised per head. The heat-kernel MLP maps the eight values of $\tilde{\mathbf{h}}_j$ through
two hidden layers of 64 units (GELU) to $d_h/2=32$ angles, each applied to a channel pair in the rotate-half convention; the
temporal encoding uses frequencies $\omega_m=700^{-2m/d_h}$.

\paragraph{Invariance of the joint coordinate.} \emph{Claim.} For any relabelling $\Pi$ of the joints and any choice of
eigenvectors, $\tilde{\mathbf{h}}$ of \cref{eq:hks} computed on the relabelled tree equals $\Pi\tilde{\mathbf{h}}$.
\emph{Proof.} A single-joint rig has $\tilde{\mathbf{h}}=\mathbf{0}$ by convention. For $J\ge2$, relabelling gives $\mathbf{A}'=\Pi\mathbf{A}\Pi^\top$ and $\mathbf{D}'=\Pi\mathbf{D}\Pi^\top$, hence
$\mathbf{L}'=\Pi\mathbf{L}\Pi^\top$ and $e^{-\tau\mathbf{L}'}=\Pi e^{-\tau\mathbf{L}}\Pi^\top$, whose diagonal is the permuted
diagonal. For an eigenvalue $\lambda$ with orthonormal eigenvectors $\mathbf{V}_\lambda\in\R^{J\times m}$,
$\sum_{k:\lambda_k=\lambda}v_{kj}^2=(\mathbf{V}_\lambda\mathbf{V}_\lambda^\top)_{jj}$ is a diagonal entry of the eigenspace's
projector, which is the same for every orthonormal basis $\mathbf{V}_\lambda\mathbf{Q}$ and every sign. Hence $h_j(\tau)=\big(e^{-\tau\mathbf{L}}\big)_{jj}-v_{0j}^2$ with the zero-eigenvalue mode $v_{0j}^2=d_j/\sum_id_i$ depends on neither the basis nor the signs, and
both terms permute with the joints; the mean over the joints is unchanged by a permutation. $\square$
Every other per-joint input is permuted with its joint, the bias $b_{ij}$ is a function of the pair, attention over joints
and per-token layers are permutation-equivariant, and the root's global channels travel with the root; so the network's
output for each joint is the same under any listing of the joints. Since all $J-1$ non-trivial modes enter \cref{eq:hks}, no truncation can split an
eigenspace.

\section{Objective and calibration}\label{app:calib}

\paragraph{Loss.} The flow time is sampled uniformly, $\sigma_{\min}=0.2$, and the Huber threshold is $\delta=10$ on normalised
residuals.
The nine groups form six families with the following target shares $s_f$ of the objective: root trajectory (root position and ground track)
$0.304$, heading $0.012$, body position $0.304$, rotation (body and root) $0.304$, velocity (body and root) $0.027$ and contact
$0.049$. Each group's energy is the mean squared normalised target over its valid entries, measured over one epoch of training
draws (55,984 windows on 5,127 rigs, batch 16, with the rest frame, crops and augmentation of training); every group $g$ of family $f$ receives the weight $\gamma_g=\gamma_f\propto\sqrt{s_f/\sum_{g'\in f}E_{g'}}$, scaled so
that the rotation family has weight 10. This gives $6.13$ for the root trajectory,
$9.95$ for body positions, $10$ for rotations, $1.22$ for velocities, $4.59$ for contact and $2.59$ for heading. Over 30 steps without optimiser updates, each family's share of
$\lVert\partial\mathcal{L}/\partial\hat\x\rVert^2$ is within a factor of $1.25$ of its target. The auxiliary weights are $0.07$ for FK consistency (ramped over the first 5k steps), $0.01$ for world velocity,
$0.01$ for foot lock and $1.0$ for acceleration; none carries the time weight.

\section{Training details}\label{app:training}

\begin{table}[!htbp]
\centering\footnotesize
\caption{Settings of \method.}\label{tab:hparams}
\setlength{\tabcolsep}{3pt}
\begin{tabular}{@{}lp{0.55\columnwidth}@{}}
\toprule
Width $D$, blocks $N_b$, heads & 512, 10, 8 (73.3M parameters) \\
Frames & 1 rest frame + $L{=}240$ target frames \\
Heat-kernel diffusion times & $\tau_i=2^{i-2}$, $i=0,\dots,7$ \\
Temporal rotary base & 700 \\
Caption dropout & 0.1 \\
Epochs, draws per epoch & 600, 55,984 (437 steps of 128; the last 48 dropped) \\
Optimiser & AdamW, weight decay 0.01 \\
Learning rate & $3{\times}10^{-4}$, 4k warm-up, half-cosine to $3{\times}10^{-6}$ \\
Gradient clip; spike rejection & 1.0; steps with gradient norm above 200 are skipped \\
Precision, hardware & bf16, 8 H200 GPUs \\
Rig balance & draw probability $\propto n^{1/2}$ \\
Skeleton augmentation & $p{=}0.8$, one of: insert a joint at 0.3--0.7 of a bone; remove 5--15\% of the leaves ($\le3$); merge 10--30\% of the single-child joints ($\ge1$); scale every bone by up to $\pm10\%$ \\
Rest-convention augmentation & $p{=}0.5$, per-joint rotation $\le30^\circ$ \\
Sampling & 50 Euler steps, guidance scale 2 \\
\bottomrule
\end{tabular}
\end{table}

\Cref{tab:hparams} lists the settings of \method. Each of the four ablation arms changes one setting and, like the full-model
control, is trained for 300 epochs with the learning-rate decay spread over 300 epochs; each arm is recalibrated (\cref{app:calib}) on its own data and model, and the control uses the main model's
calibration.

\section{Benchmark construction}\label{app:bench}

\paragraph{Cut.} Held-out units are removed from training and validation in every split of every corpus: 3,372 of the 8,610
UniML3D clips (the benchmark units, the rigs outside 5--60 joints and the clips that UniMate's own preprocessing drops), 15,960 of the 45,605 clips of the game additions
(held-out species, rigs whose clips are all animated in place, and clips without a recommended caption) and 278 of the 986 Truebones
clips (the 94 benchmark clips of the eight held-out rigs, 175 clips without source captions
and nine whose source captions all begin with ``Rest pose''). The remaining 5,238 UniML3D clips form its training (5,023)
and validation (215) splits.

\paragraph{Buckets.} OOD-tree holds out 25\% of the classes of isomorphic biped and quadruped trees that have at most three
rigs, OOD-plan 40\% of the tree classes of rare body plans; ID-clip holds out 25\% of the clips of rigs that have at least three clips (at most six per rig, keeping
at least one); ID-tree holds out 20\% of the rigs of tree classes with at least four rigs (at least
three kept per class). UniML3D test clips must move (mean joint speed of at least $0.04$ rest diagonals per second) and last
at least 20 frames. A test clip whose rig and motion are duplicated in training is removed from training. Units are assigned by hashing, one third to dev and two thirds to test, groups of geometrically identical rigs are kept
in the same split, and for the game species the species whose animations are more often shared with other species is
assigned to dev.

\paragraph{Annotations.} Tree distance is the constrained unordered tree edit distance~\cite{zhang1996constrained} (with unit costs for insertion and deletion) to the nearest training tree. A tier records whether the tree is seen (T0), new but with the same set of local patterns as a training tree (T1, Weisfeiler--Lehman patterns of height 3), or new in its local patterns (T2, subdivided into T2a and T2b at the median
pattern distance); \cref{tab:tiers} gives the tiers per bucket. Reused motion is detected on UniML3D by
fingerprints of the world directions of eight limb bones, aligned in heading and resampled to 32 frames (a near-duplicate requires a mean cosine of at least $0.98$ and a frame-difference cosine of at least
$0.95$), and on the game species by the presence of the same animation on another species, matched by action name and joint rotations. Across the benchmark, 502 clips are shorter than 30 frames and 585 are
longer than 240 frames; the latter are scored on their first 240 frames.

\begin{table}[H]
\centering\footnotesize
\caption{Topology tiers per bucket (number of clips, dev and test combined).}\label{tab:tiers}
\setlength{\tabcolsep}{4pt}
\begin{tabular}{@{}lrrrr@{}}
\toprule
Bucket & T0 & T1 & T2a & T2b \\
\midrule
ID-clip, ID-tree & 1,005 & -- & -- & -- \\
OOD-tree & -- & 169 & 145 & 219 \\
OOD-plan & -- & 8 & 32 & 62 \\
OOD-J$>$60 (Mixamo) & -- & 505 & -- & -- \\
OOD-J$>$60 (other) & -- & 356 & 120 & 202 \\
OOD-J$<$5 & -- & 1 & -- & 1 \\
OOD-species & 306 & 419 & 449 & 91 \\
OOD-new-species & 2 & -- & 10 & 36 \\
OOD-rig & -- & -- & 10 & 84 \\
\bottomrule
\end{tabular}
\end{table}

\paragraph{Captions.} Each UniML3D clip carries its single generated caption. Game-species clips carry rule-based
captions, from which a curated subset is recommended for text use; 227 OOD-species clips (103 dev, 124 test) fall outside
this subset, so they are included in the motion-quality metrics but excluded from the retrieval evaluation
(\cref{app:evaluator}). The new-species clips carry rule-based captions. Truebones clips carry captions that name the
species, in the style of AniMo4D.

\paragraph{Release.} The source assets carry their own licences. \bench will be released as per-clip source identifiers,
splits and annotations, together with the pinned UniML3D commit, the conversion scripts and the metric code.

\section{Metric definitions}\label{app:metrics}

All statistics are computed on the animation that drives the rig, at 30\,fps, with $+Y$ up and the ground at $y{=}0$; $s$ is the rest
diagonal of the rig.

\paragraph{Kinetic features and \fidk.} UniMate computes its FID on the kinetic features of
Bailando~\cite{siyao2022bailando} and Motion2Motion~\cite{chen2025motion2motion}. Both releases provide a function
\texttt{extract\_kinetic\_features} that returns three numbers per joint: the average
horizontal and vertical kinetic energy and an average acceleration term, which in Motion2Motion's release is the mean magnitude
of a window-averaged second difference of the positions and in Bailando's release
also depends on the joint's position itself, through the operator precedence in its velocity term. We run each
release unchanged, with its default arguments, on world joint positions divided by $s$ and taken relative to the first-frame
root position. A vector of $3J$ numbers cannot be compared across skeletons, so each of the three numbers is pooled over the
joints by its mean and population standard deviation: six numbers per sequence on every skeleton. Both sets are standardised
by the mean and standard deviation of the real set and compared by the Fréchet distance. \Cref{tab:main} reports \fidk{} with Motion2Motion's
release, and \cref{tab:sd:dist} also reports it with Bailando's. The
distance is computed per seed and averaged over the five seeds, for groups of at least 50 real clips.

\paragraph{Kinematic descriptors, \fidd{} and \kidk.} The descriptor of a clip comprises 24 numbers: root speed (mean,
standard deviation, 90th percentile), root
height range, absolute yaw rate (mean) and its standard deviation, root-relative joint speed (10th, 50th, 90th percentiles),
acceleration (50th, 90th), jerk (50th), mean activity, contact fraction (codec rule), airborne fraction, dominant frequency of
the activity signal, pose spread along three axes relative to the rest pose, net displacement, path length, mean root height,
median lowest-joint height and mean vertical root speed. Scale-like features are log-transformed as $\log(x+10^{-3})$; all are
standardised by the training clips' statistics and clipped at $\pm5$. \fidd is the Fréchet distance between Gaussians fitted
to the descriptors of the generated samples of a group and of its real clips, computed when the group has at least 50 real clips; it is computed per seed and then averaged in the same-data
comparison, and with all seeds pooled in the comparison with released weights (\cref{tab:released}) and on dev
(\cref{tab:cfg}). \kidk compares the clip distributions of a group per seed, with the kernel $k(x,y)=(x^\top y/24+1)^3$ and the
unbiased estimator averaged over 100 random subsets of $\min(50,\lfloor N/2\rfloor)$ clips per side, where $N$ is the number of clips in the group; its real--real value is the \kidk{} between two halves of the real clips split by rig, averaged over ten splits.

\paragraph{Skate, Slide and Cov.} We follow UniMate's evaluation of foot
sliding~\cite{mou2026unimate,karunratanakul2023gmd,zhang2018mann}. A rig's contact joints
are the joints whose canonical name contains foot, toe, ball, paw, hoof or pastern and none of whose descendants does, \ie the
distal-most such joint of each limb chain. Canonical names are given by UniMate's rule-based name cleaner, which it applies to new
assets, or, in the comparison with released weights, by its joint annotation where one exists for the rig. $\ell$ is the height of the root above the
plane through the lowest rest-pose joint. For a contact joint at frame $t$, $h$ is its height and $d$ its horizontal
displacement since frame $t-1$, both in units of $\ell$. A frame is near the ground when $h<0.05$ and skates when also
$d>0.025$. Per sequence, Skate is the share of near-ground frames that skate and Slide the mean of $100\,d\,(2-2^{h/0.05})$
over near-ground frames; both are averaged over the sequences, \ie (clip, seed) pairs, in which a contact joint comes near the
ground. Cov.\ is the share of contact-joint frames near the ground, pooled over the group. Rigs without a named contact joint,
such as snakes, fish and most insects, and rigs with $\ell=0$ are excluded from all three measures. UniMate measures Skate and Slide on 20
biped and quadruped rigs; here every rig with a named contact joint is included.

\paragraph{Ground contact and activity.} Slip is the share of grounded joint-frames, over all joints, whose horizontal
speed exceeds $0.25\,s$ per second; a joint is grounded at frame $t$ when its height is at most $0.05\,s$ at frames $t$ and
$t+1$; these are the codec's contact thresholds. Pen.\ is the share of a clip's frames whose lowest joint lies more than $0.005\,s$
below the ground. Float is the share of clips whose median lowest-joint height over frames, in units of $s$, exceeds the 95th
percentile of the same quantity over a calibration sample of training clips of the clip's body-plan group (biped, quadruped, other);
clips whose caption contains one of the words fly, flies, flying, jump, jumps, jumping, leap, leaps, hop, hops, swim, swims,
swimming, glide, glides, soar, soars, hover or hovers are excluded both from this statistic and from the
calibration sample. Speed is the root-relative joint speed
$\lVert\Delta\mathbf{p}\rVert\,\mathrm{fps}/s$, averaged over frames and joints, summed over the clips of a group with
rig-equal weights and divided by the same sum on the real clips. Pen.\ and Float are averaged within each rig and then over
rigs, as is Slip; these statistics and \kidk are computed per seed and averaged over the five seeds.

\paragraph{Self-tests.} Before use, the statistics of our scorer passed a planted-defect test on the real clips: sinking a
clip raises Pen.\ with the depth, lifting it raises the lowest-joint height by exactly the lift, time warps by $\times0.5$ and
$\times2$ scale Speed accordingly, and white noise raises \kidk monotonically with its amplitude.

\paragraph{Reproducibility.} The five seeds are 7, 13, 23, 31 and 42. Generation is deterministic per (checkpoint, clip, seed,
guidance scale): \method{} generates every sample with batch size 1 from noise seeded by a hash of the seed and the clip identifier. UniMate's
same-data test samples are solved in fixed groups of 64 cases, formed in clip order before the cases are split across GPUs
(\cref{app:baselines}).

\section{Guidance scale}\label{app:cfg}

\begin{table*}[t]
\centering\small
\caption{\method{} on all dev buckets (\nCfgRigs{} rigs, \nCfgClips{} clips, full length, five seeds per clip, 50 Euler steps) at three
guidance scales. The real-clip row gives the real values and, under \kidk, the real--real value.}\label{tab:cfg}
\begin{tabular}{@{}lcccccccc@{}}
\toprule
& \fidd$\downarrow$ & Skate$\rightarrow$ & Slide$\rightarrow$ & Cov.$\rightarrow$ & Pen.$\downarrow$ & Float$\rightarrow$ & Speed$\rightarrow$ & \kidk$\downarrow$ \\
\midrule
Real clips & -- & \nCfgRealSkate & \nCfgRealSlide & \nCfgRealCov & \nCfgRealPen & \nCfgRealFloat & 1 & \nCfgRealFloor \\
\midrule
$w{=}1.5$ & \nCfgAFID & \nCfgASkate & \nCfgASlide & \nCfgACov & \nCfgAPen & \nCfgAFloat & \nCfgASpeed & \nCfgAKID \\
$w{=}2$   & \nCfgBFID & \nCfgBSkate & \nCfgBSlide & \nCfgBCov & \nCfgBPen & \nCfgBFloat & \nCfgBSpeed & \nCfgBKID \\
$w{=}3$   & \nCfgCFID & \nCfgCSkate & \nCfgCSlide & \nCfgCCov & \nCfgCPen & \nCfgCFloat & \nCfgCSpeed & \nCfgCKID \\
\bottomrule
\end{tabular}
\end{table*}

For \method's own evaluation (\cref{tab:perbucket,tab:ablation}) and the comparison with released weights, the protocol fixes the guidance scale of each method on dev: the scale with the lowest mean \kidk{} among 1.5, 2 and 3 is selected,
with ties resolved in favour of 2, where a scale counts as tied when its gap to the lowest is within the larger of the two
95\% seed half-widths. For \method{} at 50 steps (\cref{tab:cfg}) this is 2 (\kidk{} \nCfgBKID, against \nCfgAKID{} at 1.5 and \nCfgCKID{} at 3); the ablations, sampled with 20 steps, use the scale selected at that step count,
which is also 2. For UniMate's released weights (\cref{app:released}), on the \nUcfgClips{} dev clips of the game-library
rigs at 60 frames, \kidk{} is \nUcfgAKID, \nUcfgBKID{} and \nUcfgCKID{} at 1.5, 2 and 3, so it is sampled at 1.5. AnyTop is unconditional and has no guidance. In the same-data comparison each method's setting is chosen on dev as well (\cref{app:baselines}).

\section{Comparison with UniMate}\label{app:baselines}

\paragraph{Training on the same clips.} Both methods are trained on the 5,023 clips of the UniML3D training split, each
from its own preprocessed copy of the data. UniMate is trained with its official code and released configuration for
100k steps (192 samples per
step, exponential moving average weights), with two local changes outside the model and the objective (a sampler
initialisation for our PyTorch version and an abort on a non-finite loss); it reads its released features, which hold at most the first 200 frames of a clip, and draws a random 60-frame window from them. \method{} trains as described in \cref{sec:method} for 240 epochs on
the first $\min(T,240)$ frames of each clip.

\paragraph{Sampler settings.} Each method's guidance is selected at its stated sampler configuration on the 578 UniML3D dev
clips that neither model trained on (five seeds, first $\min(60,T)$ frames) by the lowest pooled \fidk, and the test clips
are scored at that guidance. UniMate keeps
its released inference, an adaptive Dormand--Prince solver at its default tolerances, and its guidance is selected from 1.5, 2, 3, 4,
5, 7 and 10, a grid we extended until the best value lay in its interior (dev \fidk{} \nTwoDevUniOnePointFive,
\nTwoDevUniTwo, \nTwoDevUniThree, \nTwoDevUniFour, \nTwoDevUniFive, \nTwoDevUniSeven{} and \nTwoDevUniTen), so it is sampled at 7.
Selected instead by the lowest dev \fidd{} (\nThrDevUniDescOnePointFive, \nThrDevUniDescTwo, \nThrDevUniDescThree,
\nThrDevUniDescFour, \nThrDevUniDescFive, \nThrDevUniDescSeven{} and \nThrDevUniDescTen{} at the same scales), its guidance is
\nThrUniCfg; \cref{tab:main} reports both and
\cref{tab:sd:g3} gives additional metrics at \nThrUniCfg. \method{} samples with \nTwoOursSteps{} Euler steps at guidance \nTwoOursCfg, the lowest dev \fidk{} at that step count
(\nTwoDevOursTwoStepsTwoHundred; 400 steps reach \nTwoDevOursTwoStepsFourHundred); among guidance 1.5, 2 and 3, 2 also has the
lowest dev \fidd{} (\nTwoDevOursDescOnePointFive, \nTwoDevOursDescTwo{} and \nTwoDevOursDescThree). This takes 400 network
evaluations, against \nCostUniSevenNfe{} for UniMate at guidance \nTwoUniCfg{} at batch 1 and \nCostUniSevenNfeBatch{} in the
groups of 64 solved on test (\cref{tab:cost}).

\begin{table}[!htbp]
\centering\footnotesize
\caption{Network evaluations and seconds per sample on one H200 NVL (50 dev clips, seed 7, batch 1; \method: the model of \cref{tab:main}, at the clip's length). Under classifier-free guidance, each Euler step and each function evaluation of UniMate's
solver requires two network evaluations.}\label{tab:cost}
\setlength{\tabcolsep}{5pt}
\begin{tabular}{@{}lcc@{}}
\toprule
& Evaluations & Seconds \\
\midrule
\method, 20 steps  & \nCostOursClipTwentyNfe      & \nCostOursClipTwentySec \\
\method, 50 steps  & \nCostOursClipFiftyNfe       & \nCostOursClipFiftySec \\
\method, 100 steps & \nCostOursClipHundredNfe     & \nCostOursClipHundredSec \\
\method, 200 steps & \nCostOursClipTwoHundredNfe  & \nCostOursClipTwoHundredSec \\
\method, 400 steps & \nCostOursClipFourHundredNfe & \nCostOursClipFourHundredSec \\
\midrule
UniMate, guidance 3 & \nCostUniThreeNfe & \nCostUniThreeSec \\
UniMate, guidance 4 & \nCostUniFourNfe  & \nCostUniFourSec \\
UniMate, guidance 5 & \nCostUniFiveNfe  & \nCostUniFiveSec \\
UniMate, guidance 7 & \nCostUniSevenNfe & \nCostUniSevenSec \\
\bottomrule
\end{tabular}
\end{table}

\paragraph{Scoring.} Every sequence is scored on its first $\min(60,T)$ frames against the first $\min(60,T)$ frames of the real clips. UniMate's test cases are solved in fixed groups of 64, within
which the adaptive step is shared. Differences between the methods are assessed by a paired bootstrap over rigs (2,000 resamples, drawing all clips
and seeds of a rig together) and deemed resolved when the 95\% interval excludes zero; groups in which the difference is
undefined for some resamples are not evaluated. \fidd's real--real value is the mean, over 200 random splits, of the distance
between the two halves of the real clips. \Cref{tab:sd:dist,tab:sd:contact} report selected metrics for every group with UniMate tuned on \fidk; \cref{tab:sd:g3}
reports selected metrics with UniMate tuned on \fidd. 

\begin{table*}[t]
\centering\scriptsize
\caption{Same-data comparison, Fréchet distances (test, first 60 frames, five seeds; mean over seeds). Ours: \nTwoOursSteps{}
Euler steps at guidance \nTwoOursCfg, at the clip's length; UniMate: guidance \nTwoUniCfg. \fidk{} on the Motion2Motion and the Bailando release of the kinetic
features. Bold: the lower value, where the paired 95\% confidence interval over rigs excludes zero. R--R: real clips, half against half (groups of at least 100 clips).}\label{tab:sd:dist}
\setlength{\tabcolsep}{3pt}
\begin{tabular}{@{}lr|ccc|cc|cc@{}}
\toprule
& & \multicolumn{3}{c|}{\fidd} & \multicolumn{2}{c|}{\fidk, Motion2Motion} & \multicolumn{2}{c}{\fidk, Bailando} \\
Test clips & Clips & Ours & UniMate & R--R & Ours & UniMate & Ours & UniMate \\
\midrule
All & \nTwoAllClips & \nTwoAllDescOursB & \nTwoAllDescUniB & \nTwoAllDescFloor & \nTwoAllKinOursB & \nTwoAllKinUniB & \nTwoAllBkOursB & \nTwoAllBkUniB \\
ID-clip & \nTwoClipClips & \nTwoClipDescOursB & \nTwoClipDescUniB & \nTwoClipDescFloor & \nTwoClipKinOursB & \nTwoClipKinUniB & \nTwoClipBkOursB & \nTwoClipBkUniB \\
No training clip & \nTwoNoTrClips & \nTwoNoTrDescOursB & \nTwoNoTrDescUniB & \nTwoNoTrDescFloor & \nTwoNoTrKinOursB & \nTwoNoTrKinUniB & \nTwoNoTrBkOursB & \nTwoNoTrBkUniB \\
\quad novel motion & \nTwoNovelClips & \nTwoNovelDescOursB & \nTwoNovelDescUniB & \nTwoNovelDescFloor & \nTwoNovelKinOursB & \nTwoNovelKinUniB & \nTwoNovelBkOursB & \nTwoNovelBkUniB \\
\quad reused motion & \nTwoReusedClips & \nTwoReusedDescOursB & \nTwoReusedDescUniB & \nTwoReusedDescFloor & \nTwoReusedKinOursB & \nTwoReusedKinUniB & \nTwoReusedBkOursB & \nTwoReusedBkUniB \\
\quad ID-tree & \nTwoSTreeClips & \nTwoSTreeDescOursB & \nTwoSTreeDescUniB & \nTwoSTreeDescFloor & \nTwoSTreeKinOursB & \nTwoSTreeKinUniB & \nTwoSTreeBkOursB & \nTwoSTreeBkUniB \\
\quad OOD-tree & \nTwoTreeClips & \nTwoTreeDescOursB & \nTwoTreeDescUniB & \nTwoTreeDescFloor & \nTwoTreeKinOursB & \nTwoTreeKinUniB & \nTwoTreeBkOursB & \nTwoTreeBkUniB \\
\quad OOD-plan & \nTwoPlanClips & \nTwoPlanDescOursB & \nTwoPlanDescUniB & -- & \nTwoPlanKinOursB & \nTwoPlanKinUniB & \nTwoPlanBkOursB & \nTwoPlanBkUniB \\
\bottomrule
\end{tabular}
\end{table*}

\begin{table*}[t]
\centering\scriptsize
\caption{Same-data comparison, ground contact and activity (test, first 60 frames, five seeds; Ours and UniMate as in
\cref{tab:sd:dist}). Pen.: share of frames below the ground (0 for real clips). Skate and Cov.: closer to the real value is better. Speed: joint speed relative to the real clips. Bold: lower Pen., or a smaller absolute difference from the real value for Skate and Cov., where the paired 95\% confidence interval over rigs excludes zero.}\label{tab:sd:contact}
\setlength{\tabcolsep}{3.5pt}
\begin{tabular}{@{}l|cc|ccc|ccc|cc@{}}
\toprule
& \multicolumn{2}{c|}{Pen.$\downarrow$} & \multicolumn{3}{c|}{Skate} & \multicolumn{3}{c|}{Cov.}
& \multicolumn{2}{c}{Speed} \\
Test clips & Ours & UniMate & Ours & UniMate & Real & Ours & UniMate & Real & Ours & UniMate \\
\midrule
All & \nTwoAllPenOursB & \nTwoAllPenUniB & \nTwoAllSkateOursB & \nTwoAllSkateUniB & \nTwoAllSkateReal & \nTwoAllCovOursB & \nTwoAllCovUniB & \nTwoAllCovReal & \nTwoAllSpeedOurs & \nTwoAllSpeedUni \\
ID-clip & \nTwoClipPenOursB & \nTwoClipPenUniB & \nTwoClipSkateOursB & \nTwoClipSkateUniB & \nTwoClipSkateReal & \nTwoClipCovOursB & \nTwoClipCovUniB & \nTwoClipCovReal & \nTwoClipSpeedOurs & \nTwoClipSpeedUni \\
No training clip & \nTwoNoTrPenOursB & \nTwoNoTrPenUniB & \nTwoNoTrSkateOursB & \nTwoNoTrSkateUniB & \nTwoNoTrSkateReal & \nTwoNoTrCovOursB & \nTwoNoTrCovUniB & \nTwoNoTrCovReal & \nTwoNoTrSpeedOurs & \nTwoNoTrSpeedUni \\
\quad novel motion & \nTwoNovelPenOursB & \nTwoNovelPenUniB & \nTwoNovelSkateOursB & \nTwoNovelSkateUniB & \nTwoNovelSkateReal & \nTwoNovelCovOursB & \nTwoNovelCovUniB & \nTwoNovelCovReal & \nTwoNovelSpeedOurs & \nTwoNovelSpeedUni \\
\quad reused motion & \nTwoReusedPenOursB & \nTwoReusedPenUniB & \nTwoReusedSkateOursB & \nTwoReusedSkateUniB & \nTwoReusedSkateReal & \nTwoReusedCovOursB & \nTwoReusedCovUniB & \nTwoReusedCovReal & \nTwoReusedSpeedOurs & \nTwoReusedSpeedUni \\
\quad ID-tree & \nTwoSTreePenOursB & \nTwoSTreePenUniB & \nTwoSTreeSkateOursB & \nTwoSTreeSkateUniB & \nTwoSTreeSkateReal & \nTwoSTreeCovOursB & \nTwoSTreeCovUniB & \nTwoSTreeCovReal & \nTwoSTreeSpeedOurs & \nTwoSTreeSpeedUni \\
\quad OOD-tree & \nTwoTreePenOursB & \nTwoTreePenUniB & \nTwoTreeSkateOursB & \nTwoTreeSkateUniB & \nTwoTreeSkateReal & \nTwoTreeCovOursB & \nTwoTreeCovUniB & \nTwoTreeCovReal & \nTwoTreeSpeedOurs & \nTwoTreeSpeedUni \\
\quad OOD-plan & \nTwoPlanPenOursB & \nTwoPlanPenUniB & \nTwoPlanSkateOursB & \nTwoPlanSkateUniB & \nTwoPlanSkateReal & \nTwoPlanCovOursB & \nTwoPlanCovUniB & \nTwoPlanCovReal & \nTwoPlanSpeedOurs & \nTwoPlanSpeedUni \\
\bottomrule
\end{tabular}
\end{table*}

\begin{table*}[t]
\centering\scriptsize
\caption{Same-data comparison with UniMate tuned on dev \fidd{} (guidance \nThrUniCfg; test, first 60 frames, five seeds;
Ours as in \cref{tab:sd:dist}). \fidk: Motion2Motion (M2M) and Bailando (Bk.) releases of the kinetic features. Skate: closer to the real value
is better; Speed: relative to the real clips (target 1). Bold: lower \fidd, \fidk{} or Pen., or closer
to the real value for Skate and Speed, where the paired 95\% confidence interval over rigs excludes zero.}\label{tab:sd:g3}
\setlength{\tabcolsep}{3pt}
\begin{tabular}{@{}lr|cc|cc|cc|cc|ccc|cc@{}}
\toprule
& & \multicolumn{2}{c|}{\fidd$\downarrow$} & \multicolumn{2}{c|}{\fidk{} M2M$\downarrow$} & \multicolumn{2}{c|}{\fidk{} Bk.$\downarrow$}
& \multicolumn{2}{c|}{Pen.$\downarrow$} & \multicolumn{3}{c|}{Skate} & \multicolumn{2}{c}{Speed} \\
Test clips & Clips & Ours & Uni. & Ours & Uni. & Ours & Uni. & Ours & Uni. & Ours & Uni. & Real & Ours & Uni. \\
\midrule
All                    & \nThrAllClips & \nThrAllDescOursB & \nThrAllDescUniB & \nThrAllKinOursB & \nThrAllKinUniB & \nThrAllBkOursB & \nThrAllBkUniB & \nThrAllPenOursB & \nThrAllPenUniB & \nThrAllSkateOursB & \nThrAllSkateUniB & \nThrAllSkateReal & \nThrAllSpeedOursB & \nThrAllSpeedUniB \\
ID-clip                & \nThrClipClips & \nThrClipDescOursB & \nThrClipDescUniB & \nThrClipKinOursB & \nThrClipKinUniB & \nThrClipBkOursB & \nThrClipBkUniB & \nThrClipPenOursB & \nThrClipPenUniB & \nThrClipSkateOursB & \nThrClipSkateUniB & \nThrClipSkateReal & \nThrClipSpeedOursB & \nThrClipSpeedUniB \\
No training clip       & \nThrNoTrClips & \nThrNoTrDescOursB & \nThrNoTrDescUniB & \nThrNoTrKinOursB & \nThrNoTrKinUniB & \nThrNoTrBkOursB & \nThrNoTrBkUniB & \nThrNoTrPenOursB & \nThrNoTrPenUniB & \nThrNoTrSkateOursB & \nThrNoTrSkateUniB & \nThrNoTrSkateReal & \nThrNoTrSpeedOursB & \nThrNoTrSpeedUniB \\
\quad novel motion     & \nThrNovelClips & \nThrNovelDescOursB & \nThrNovelDescUniB & \nThrNovelKinOursB & \nThrNovelKinUniB & \nThrNovelBkOursB & \nThrNovelBkUniB & \nThrNovelPenOursB & \nThrNovelPenUniB & \nThrNovelSkateOursB & \nThrNovelSkateUniB & \nThrNovelSkateReal & \nThrNovelSpeedOursB & \nThrNovelSpeedUniB \\
\quad reused motion    & \nThrReusedClips & \nThrReusedDescOursB & \nThrReusedDescUniB & \nThrReusedKinOursB & \nThrReusedKinUniB & \nThrReusedBkOursB & \nThrReusedBkUniB & \nThrReusedPenOursB & \nThrReusedPenUniB & \nThrReusedSkateOursB & \nThrReusedSkateUniB & \nThrReusedSkateReal & \nThrReusedSpeedOursB & \nThrReusedSpeedUniB \\
\quad ID-tree          & \nThrSTreeClips & \nThrSTreeDescOursB & \nThrSTreeDescUniB & \nThrSTreeKinOursB & \nThrSTreeKinUniB & \nThrSTreeBkOursB & \nThrSTreeBkUniB & \nThrSTreePenOursB & \nThrSTreePenUniB & \nThrSTreeSkateOursB & \nThrSTreeSkateUniB & \nThrSTreeSkateReal & \nThrSTreeSpeedOursB & \nThrSTreeSpeedUniB \\
\quad OOD-tree         & \nThrTreeClips & \nThrTreeDescOursB & \nThrTreeDescUniB & \nThrTreeKinOursB & \nThrTreeKinUniB & \nThrTreeBkOursB & \nThrTreeBkUniB & \nThrTreePenOursB & \nThrTreePenUniB & \nThrTreeSkateOursB & \nThrTreeSkateUniB & \nThrTreeSkateReal & \nThrTreeSpeedOursB & \nThrTreeSpeedUniB \\
\quad OOD-plan         & \nThrPlanClips & \nThrPlanDescOursB & \nThrPlanDescUniB & \nThrPlanKinOursB & \nThrPlanKinUniB & \nThrPlanBkOursB & \nThrPlanBkUniB & \nThrPlanPenOursB & \nThrPlanPenUniB & \nThrPlanSkateOursB & \nThrPlanSkateUniB & \nThrPlanSkateReal & \nThrPlanSpeedOursB & \nThrPlanSpeedUniB \\
\bottomrule
\end{tabular}
\end{table*}

\subsection{Released weights}\label{app:released}

\paragraph{UniMate.} We use the released v2 weights (100k steps, CC BY-NC), trained on UniML3D at commit
\texttt{faaa817}, on Truebones, Mixamo and Objaverse skeletons of 5--70 joints without a held-out split; the model
generates at most 71 joints and 60 frames at 30\,fps. Each benchmark rig is converted to UniMate's T-pose input fields; UniML3D and Truebones rigs use its released joint annotations,
game rigs its rule-based name cleaning and facing joints. Its topology conditioner and ODE sampler run on one case at a time
(batch 1) at guidance scale 1.5, selected on dev (\cref{app:cfg}). Statistics for de-normalisation are chosen by domain (its
Objaverse statistics for UniML3D rigs, its Truebones statistics for animals). The FK of its rotations is mapped back to our
frame by inverting its canonicalisation; a round trip of real clips through the adapter is exact to $2.5\times10^{-7}\,s$,
and the adapter matches its official sampling script to $10^{-7}$.

\paragraph{AnyTop.} We use the released unified model, trained on 70 Truebones characters; it is not conditioned on
text. Each rig is exported to
BVH with two rest frames and its benchmark clips of the same split, from which AnyTop's own preprocessing computes the
normalisation statistics; the four facing joints it needs are the mirrored limb roots of the rest pose. It is sampled with its 100-step sampler at the length of the real clip, and its output is exported to BVH
through its official inverse kinematics; the FK of this BVH is scored.

\paragraph{Coverage.} Training membership is reconstructed per rig: for UniMate from the export lists, skip lists and joint
counts released with its data, for AnyTop from its released condition file.
On test, 842 clips are on rigs unseen by all three methods; UniMate cannot generate the 54 OOD-J$>$60 and 12 new-species
test clips with more than 71 joints, nor the OOD-J$<$5 clip. All three methods were run on the \nSpClips{} OOD-species clips and the \nTbClips{} clips of the Truebones test rigs.

\begin{table}[H]
\centering\footnotesize
\caption{Comparison with released weights (test, first 60 frames, five seeds per clip). \method{} is the main model, trained on four corpora.
Top: the two held-out species, on rigs that none of the three methods was trained on; bottom: the Truebones test rigs, on
which both baselines were trained. $^\dagger$Unconditional and transductive. Real clips give the reference values.}\label{tab:released}
\setlength{\tabcolsep}{4pt}
\begin{tabular}{@{}lcccc@{}}
\toprule
Method & \fidd$\downarrow$ & Skate$\rightarrow$ & Slide$\rightarrow$ & Cov.$\rightarrow$ \\
\midrule
\multicolumn{5}{@{}l}{\emph{Rigs unseen by all three: capybara and emu, \nSpRigs{} rigs, \nSpClips{} clips}} \\
Real clips                            & --          & \nSpRealSkate & \nSpRealSlide & \nSpRealCov \\
AnyTop$^\dagger$~\cite{gat2025anytop} & \nSpAnyFID  & \nSpAnySkate  & \nSpAnySlide  & \nSpAnyCov \\
UniMate~\cite{mou2026unimate}         & \nSpUniFID  & \nSpUniSkate  & \nSpUniSlide  & \nSpUniCov \\
\method                               & \nSpOursFID & \nSpOursSkate & \nSpOursSlide & \nSpOursCov \\
\midrule
\multicolumn{5}{@{}l}{\emph{Trained on by both baselines: \nTbRigs{} Truebones rigs, \nTbClips{} clips}} \\
Real clips                            & --          & \nTbRealSkate & \nTbRealSlide & \nTbRealCov \\
AnyTop$^\dagger$                      & \nTbAnyFID  & \nTbAnySkate  & \nTbAnySlide  & \nTbAnyCov \\
UniMate                               & \nTbUniFID  & \nTbUniSkate  & \nTbUniSlide  & \nTbUniCov \\
\method                               & \nTbOursFID & \nTbOursSkate & \nTbOursSlide & \nTbOursCov \\
\bottomrule
\end{tabular}
\end{table}

\paragraph{Results.} On the two held-out species, whose rigs none of the three methods was trained on, \method{} attains a lower \fidd{} (\nSpOursFID) than UniMate (\nSpUniFID) and AnyTop (\nSpAnyFID), and its skating and sliding are closest to the real clips (Skate \nSpOursSkate, compared with \nSpUniSkate{} and \nSpAnySkate; \nSpRealSkate{} for real clips). On the Truebones test rigs, on which both
baselines were trained, the ranking by \fidd{} is the same (\nTbOursFID, \nTbUniFID{} and \nTbAnyFID).

\section{Ablations per bucket}\label{app:abl}

\begin{table*}[t]
\centering\small
\caption{\Cref{tab:ablation} with its uncertainty: \kidk{} as the mean over the five sampling seeds $\pm$ its 95\% half-width
over them. Columns: full model, learned frame table, learned joint table, eigenvectors + SignNet, without augmentation.
$^\ast$Fewer than 20 rigs.}\label{tab:abl:buckets}
\setlength{\tabcolsep}{5pt}
\begin{tabular}{@{}lrcccccc@{}}
\toprule
Bucket & Rigs / clips & Real--real & Full & Frame table & Joint table & Eigvec.\,+\,SignNet & $-$\,aug. \\
\midrule
OOD-tree               & \nAbTreeRigs{} / \nAbTreeClips       & \nAbTreeFloor    & \nAbCtlTreeKID{\,$\pm$\nAbCtlTreeKIDhw}       & \nAbTroTreeKID{\,$\pm$\nAbTroTreeKIDhw}       & \nAbJtabTreeKID{\,$\pm$\nAbJtabTreeKIDhw}       & \nAbEigTreeKID{\,$\pm$\nAbEigTreeKIDhw}       & \nAbAugTreeKID{\,$\pm$\nAbAugTreeKIDhw} \\
OOD-plan               & \nAbPlanRigs{} / \nAbPlanClips       & \nAbPlanFloor    & \nAbCtlPlanKID{\,$\pm$\nAbCtlPlanKIDhw}       & \nAbTroPlanKID{\,$\pm$\nAbTroPlanKIDhw}       & \nAbJtabPlanKID{\,$\pm$\nAbJtabPlanKIDhw}       & \nAbEigPlanKID{\,$\pm$\nAbEigPlanKIDhw}       & \nAbAugPlanKID{\,$\pm$\nAbAugPlanKIDhw} \\
OOD-new-species$^\ast$ & \nAbNewSpRigs{} / \nAbNewSpClips     & \nAbNewSpFloor   & \nAbCtlNewSpKID{\,$\pm$\nAbCtlNewSpKIDhw}     & \nAbTroNewSpKID{\,$\pm$\nAbTroNewSpKIDhw}     & \nAbJtabNewSpKID{\,$\pm$\nAbJtabNewSpKIDhw}     & \nAbEigNewSpKID{\,$\pm$\nAbEigNewSpKIDhw}     & \nAbAugNewSpKID{\,$\pm$\nAbAugNewSpKIDhw} \\
OOD-rig$^\ast$         & \nAbRigRigs{} / \nAbRigClips         & \nAbRigFloor     & \nAbCtlRigKID{\,$\pm$\nAbCtlRigKIDhw}         & \nAbTroRigKID{\,$\pm$\nAbTroRigKIDhw}         & \nAbJtabRigKID{\,$\pm$\nAbJtabRigKIDhw}         & \nAbEigRigKID{\,$\pm$\nAbEigRigKIDhw}         & \nAbAugRigKID{\,$\pm$\nAbAugRigKIDhw} \\
\bottomrule
\end{tabular}
\end{table*}

\Cref{tab:abl:buckets} gives the seed half-widths of \cref{tab:ablation}. Skeleton augmentation targets unseen topologies and body plans, which these buckets test. We test rest-convention augmentation directly on \nRcClips{} OOD test clips on \nRcRigs{} rigs: each joint's rest orientation of the
test skeleton is perturbed by its own random rotation of at most \nRcDeg$^\circ$ (the training range), the world motion is unchanged, and the model is sampled again with the same seed and caption. The median change of the generated joint positions is
\nRcCtl{} mean bone lengths with the full model and \nRcAug{} without augmentation, a reduction of \nRcDiff{} (95\% bootstrap
interval \nRcLo--\nRcHi{} over clips, \nRcRigLo--\nRcRigHi{} over rigs). For reference, resampling the same clip with a different seed changes the output by \nRcSeedCtl{} (\nRcSeedAug{}
without augmentation). The full model moves less in each of the \nRcBuckets{} OOD buckets with probe
clips.

\section{Qualitative figures}\label{app:qual}

\begin{figure*}[t]
\centering
\includegraphics[width=\textwidth]{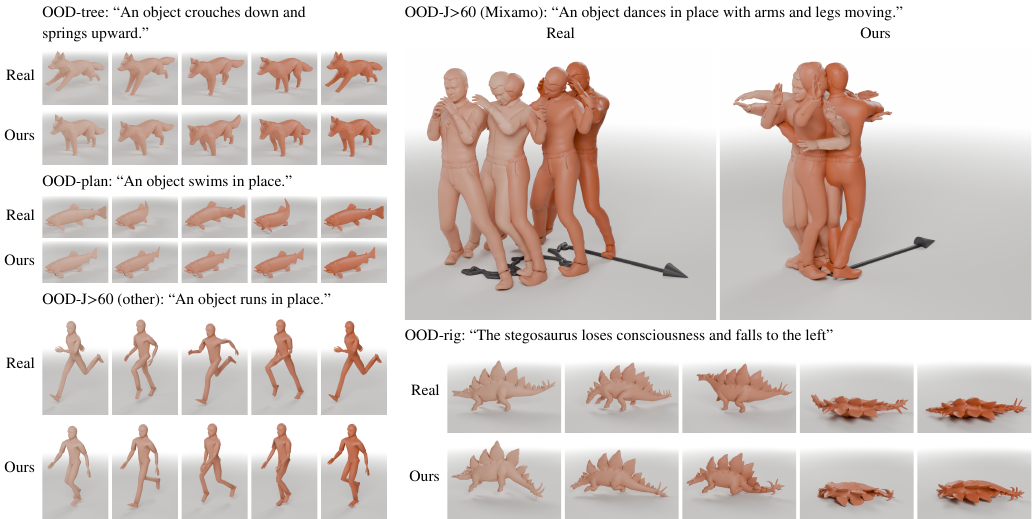}
\caption{One test clip from each of five OOD buckets, showing the real clip and the output of \method{} (guidance 2,
seed 7). Each strip shows the real clip above ours at matching times; the strobe panel (Mixamo) shows the real clip on the left and overlays five poses spaced along the real
clip's root path. Headers give the bucket and the caption.}\label{fig:qual:buckets}
\end{figure*}

\begin{figure*}[t]
\centering
\includegraphics[width=\textwidth]{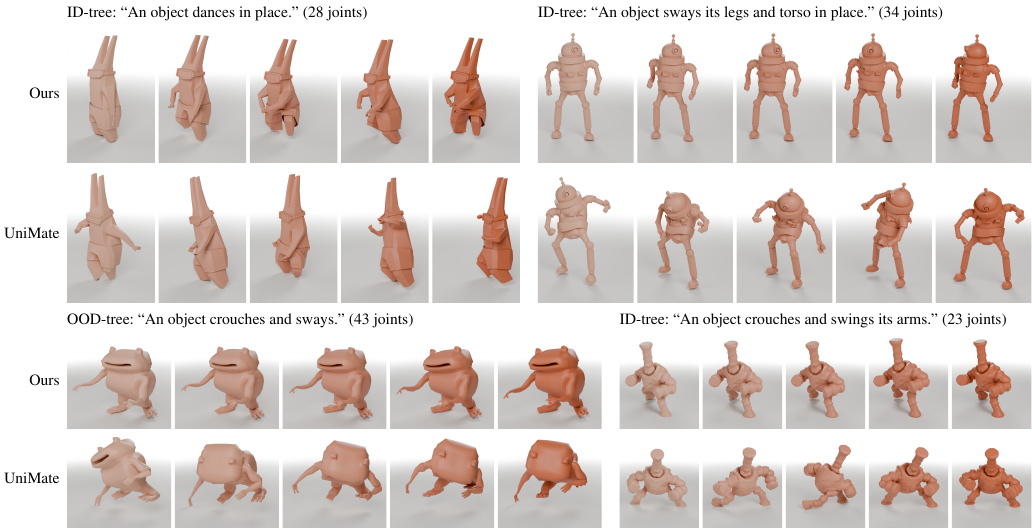}
\caption{Four additional humanoid rigs without training clips, drawn as in \cref{fig:qual}(a), with \method{} above
UniMate.}\label{fig:qual:humanoids}
\end{figure*}

\paragraph{Selection and drawing of Figs.~\ref{fig:qual}, \ref{fig:qual:buckets} and~\ref{fig:qual:humanoids}.} The examples are selected among the test clips whose rig has a rigged mesh, either
under a public licence (\cref{tab:credits}) or from the game library, that the skeleton drives faithfully. \Cref{fig:qual}(a) and Fig.~\ref{fig:qual:humanoids}
show the samples of \cref{tab:main} (UniMate tuned on \fidk, seed 7) over the first 60 frames, or the whole clip if shorter; \cref{fig:qual}(b) and \cref{fig:qual:buckets} show the main
model (50 steps, guidance 2, seed 7), \cref{fig:arch} the main model with 20 steps at guidance 2 (seed 7). The generated joint rotations drive the asset's own armature, and the mesh deforms by linear
blend skinning with the asset's own weights. All rows of a clip share one camera, one set of frames and one horizontal ground plane, placed at the data's ground height at the rig's origin and lowered by
the depth by which the rest mesh extends below the lowest joint. The Mixamo
panel of Fig.~\ref{fig:qual:buckets} overlays five poses in one image and shows the root path on the floor.

\begin{table}[H]
\centering\footnotesize
\caption{Meshes under public licences in Figs.~\ref{fig:arch}, \ref{fig:qual}, \ref{fig:qual:buckets} and~\ref{fig:qual:humanoids}; the human mesh in
\cref{fig:arch} is SMPL+H~\cite{romero2017mano}. $^\dagger$Title shortened.}\label{tab:credits}
\setlength{\tabcolsep}{3pt}
\begin{tabular}{@{}lll@{}}
\toprule
Mesh (author) & Licence & Figure \\
\midrule
\href{https://sketchfab.com/3d-models/fac0adfd62e2409cbaf9fec93d547d99}{Defeat Idle} (vicente betoret ferrero) & CC BY 4.0 & \ref{fig:qual}(a) \\
\href{https://sketchfab.com/3d-models/66843ae01ed74b50bb1613cc05758037}{Lowpoly Dog Walk} (timesen) & CC BY 4.0 & \ref{fig:qual}(a) \\
\href{https://sketchfab.com/3d-models/2355ddc0a1b34c9b84666924843ab4fb}{Dey Flair Dancing} (jdey123) & CC BY 4.0 & \ref{fig:qual}(a) \\
\href{https://sketchfab.com/3d-models/1d0b5b7ff2d94aa2bf3ae526b4253acc}{Back Flip To Uppercut (1)} (marcolinadesign) & CC BY 4.0 & \ref{fig:qual}(a) \\
\href{https://sketchfab.com/3d-models/d3ea521c69a34307b0c149864031520c}{Low poly fox running animation} (dragonsnap) & CC BY 4.0 & \ref{fig:qual:buckets} \\
\href{https://sketchfab.com/3d-models/44d4c3e01dc7479c8225ddfab3dfc8a6}{fario3} (masterdef) & CC BY 4.0 & \ref{fig:qual:buckets} \\
\href{https://sketchfab.com/3d-models/a7f426205a0f4fc7ad187f5d500c7a2c}{Human Run Cycle} (zhuqiann) & CC BY 4.0 & \ref{fig:qual:buckets} \\
\href{https://sketchfab.com/3d-models/b549280b41334c3494c66dfa5114012a}{Project Artificial Intelligence} (Headsketch) & CC BY 4.0 & \ref{fig:qual:buckets} \\
\href{https://truebones.gumroad.com/}{Stego, Spider} (Truebones) & Truebones terms & \ref{fig:arch}, \ref{fig:qual:buckets} \\
\href{https://sketchfab.com/3d-models/a06d60f0ab144adc982cdc94bf24e368}{Gangnam Style Dancing Rabbit$^\dagger$} (antonmoek) & CC BY 4.0 & \ref{fig:qual:humanoids} \\
\href{https://sketchfab.com/3d-models/854155155df749318b9110bb70f6c25d}{Looking Around} (Artur Coelho) & CC BY 4.0 & \ref{fig:qual:humanoids} \\
\href{https://sketchfab.com/3d-models/81b11403177242ea9d1242dbfe214432}{Kooku$^\dagger$} (ABCzezeze) & CC BY 4.0 & \ref{fig:qual:humanoids} \\
\href{https://sketchfab.com/3d-models/387b5e0e2c7d4041bd46380eb2d63c4a}{Fluid Guy} (Jimmy Gunawan) & CC BY 4.0 & \ref{fig:qual:humanoids} \\
\bottomrule
\end{tabular}
\end{table}

\section{Text--motion alignment}\label{app:evaluator}

\paragraph{Evaluator.} A two-tower model maps captions and motions to a shared 512-d space: a frozen DistilBERT~\cite{sanh2019distilbert}
with a trainable head, and a graph--temporal motion encoder over world joint positions (height kept, horizontal positions
relative to the smoothed root), velocities and the smoothed root's displacement, with per-joint rest geometry and the tree as
static input and no joint names; joint order is canonicalised, so a relabelling leaves the embedding bit-identical. It is
trained with a symmetric InfoNCE loss on 132,368 training clips, with 6,949
validation clips held out for model selection, with FK-exact
skeleton augmentation and species words masked in half of the captions.

\paragraph{Validity on the training distribution.} With captions de-duplicated within pools of 32 (chance 0.031) and species
words masked, captions retrieve their real held-out clips with R@1 0.92 on AniMo4D animals and 0.93 on the game-library additions; shuffling the frames removes 79\% of the margin above chance; subdividing bones, deleting leaves and rescaling
bones keep the embedding at a mean cosine of at least 0.988. Two training seeds rank 16 conditions (the real clips, 11 degradations and 4 nuisance transformations) with Kendall
$\tau=0.90$. 

\paragraph{Held-out species.} The held-out-species evaluation of \cref{sec:bench:metrics} uses the \nRetClips{} test clips
on \nRetRigs{} rigs that carry a curated caption and 30--240 frames, scored at full length; \cref{tab:textalign} lists its
results.

\paragraph{Against UniMate.} With the same evaluator and pools, we also score UniMate's released weights (guidance 1.5,
selected on dev; \cref{app:cfg}) and the same-data pair of \cref{tab:main} (UniMate tuned on \fidk), with every system
additionally scored on the first 60 frames (\cref{tab:textalign}). Retrieval always uses the curated captions; UniMate's models and the same-data \method{}
generate from these captions rewritten in UniMate's training style (``An object \dots''), and the main model generates
from the curated captions directly. Since the retrained UniMate handles at most 61 joints, the same-data pair is compared on the
\nTaEmuClips{} clips of the \nTaEmuRigs{} emu rigs; the capybara rigs have 66 joints.

\begin{table}[H]
\centering\scriptsize
\caption{Text-to-motion R@1 within rigs of the held-out game species (pools of 8, chance 0.125): mean and 95\%
half-width over five seeds, and the R@1 with captions shuffled (Shuf.). All: \nTaAllClips{} clips on \nTaAllRigs{} rigs;
emu: \nTaEmuClips{} clips on \nTaEmuRigs{} rigs.}\label{tab:textalign}
\setlength{\tabcolsep}{2pt}
\begin{tabular}{@{}lcccc@{}}
\toprule
& \multicolumn{2}{c}{All} & \multicolumn{2}{c}{Emu} \\
& R@1 & Shuf. & R@1 & Shuf. \\
\midrule
Real clips, first 60 frames & \nTaRealSixtyAllR & -- & \nTaRealSixtyEmuR & -- \\
Real clips, full length & \nTaRealFullAllR & -- & \nTaRealFullEmuR & -- \\
\midrule
\method, generating 60 frames & \nTaMainSixtyAllR$\pm$\nTaMainSixtyAllRHw & \nTaMainSixtyAllShuf & \nTaMainSixtyEmuR$\pm$\nTaMainSixtyEmuRHw & \nTaMainSixtyEmuShuf \\
\method, clip length, first 60 frames & \nTaMainClipSixtyAllR$\pm$\nTaMainClipSixtyAllRHw & \nTaMainClipSixtyAllShuf & \nTaMainClipSixtyEmuR$\pm$\nTaMainClipSixtyEmuRHw & \nTaMainClipSixtyEmuShuf \\
\method, clip length, full & \nTaMainClipFullAllR$\pm$\nTaMainClipFullAllRHw & \nTaMainClipFullAllShuf & \nTaMainClipFullEmuR$\pm$\nTaMainClipFullEmuRHw & \nTaMainClipFullEmuShuf \\
UniMate, released weights & \nTaUniOffAllR$\pm$\nTaUniOffAllRHw & \nTaUniOffAllShuf & \nTaUniOffEmuR$\pm$\nTaUniOffEmuRHw & \nTaUniOffEmuShuf \\
\midrule
Same data: \method, 200 steps, 60 frames & \nTaSameAllR$\pm$\nTaSameAllRHw & \nTaSameAllShuf & \nTaSameEmuR$\pm$\nTaSameEmuRHw & \nTaSameEmuShuf \\

Same data: UniMate, guidance \nTwoUniCfg & -- & -- & \nTaUniReEmuR$\pm$\nTaUniReEmuRHw & \nTaUniReEmuShuf \\
\bottomrule
\end{tabular}
\end{table}

\end{document}